\documentclass[twoside,11pt]{article}

\usepackage{jmlr2e}
\usepackage{multirow}
\usepackage{amssymb, amsmath}  
\usepackage{color}
\usepackage{url}
\usepackage{algorithm,algcompatible}
\algnewcommand\INPUT{\item[\textbf{Input:}]}%
\algnewcommand\OUTPUT{\item[\textbf{Output:}]}%

\usepackage{subfig}
\newcommand{\dataset}{{\cal D}}

\allowdisplaybreaks

\jmlrheading{XX}{2018}{1-44}{07/2018}{XX/XX}{Chiwoo Park and Daniel Apley}

\ShortHeadings{Patchwork Kriging}{Park and Apley}
\firstpageno{1}

\newcommand{\qed}{\nobreak \ifvmode \relax \else
      \ifdim\lastskip<1.5em \hskip-\lastskip
      \hskip1.5em plus0em minus0.5em \fi \nobreak
      \vrule height0.75em width0.5em depth0.25em\fi}

\newcommand{\M}[1]{\boldsymbol{#1}}  
\newcommand{\V}[1]{\boldsymbol{#1}}  
\newcommand{\EX}[0]{\mathbb{E}}
\newcommand{\X}[1]{\mathbf{#1}}  

\DeclareMathOperator*{\CV}{\mathbb{C}ov}
\DeclareMathOperator*{\VA}{\mathbb{V}ar}

\newtheorem{assumption}{Assumption}

\begin{document}

\title{Patchwork Kriging for Large-scale Gaussian Process Regression}

\author{\name Chiwoo Park \email cpark5@fsu.edu\\
       \addr Department of Industrial and Manufacturing Engineering\\
       Florida State University\\
       2525 Pottsdamer St., Tallahassee, FL 32310-6046, USA.
       \AND
       \name Daniel Apley \email apley@northwestern.edu \\
       \addr Dept. of Industrial Engineering and Management Sciences \\
       Northwestern University\\
      2145 Sheridan Rd., Evanston, IL 60208-3119, USA.}

\editor{Neil Lawrence}

\maketitle

\begin{abstract}
This paper presents a new approach for Gaussian process (GP) regression for large datasets. The approach involves partitioning the regression input domain into multiple local regions with a different local GP model fitted in each region. Unlike existing local partitioned GP approaches, we introduce a technique for patching together the local GP models nearly seamlessly to ensure that the local GP models for two neighboring regions produce nearly the same response prediction and prediction error variance on the boundary between the two regions. This largely mitigates the well-known discontinuity problem that degrades the prediction accuracy of existing local partitioned GP methods over regional boundaries. Our main innovation is to represent the continuity conditions as additional pseudo-observations that the differences between neighboring GP responses are identically zero at an appropriately chosen set of boundary input locations. To predict the response at any input location, we simply augment the actual response observations with the pseudo-observations and apply standard GP prediction methods to the augmented data. In contrast to heuristic continuity adjustments, this has an advantage of working within a formal GP framework, so that the GP-based predictive uncertainty quantification remains valid. Our approach also inherits a sparse block-like structure for the sample covariance matrix, which results in computationally efficient closed-form expressions for the predictive mean and variance. In addition, we provide a new spatial partitioning scheme based on a recursive space partitioning along local principal component directions, which makes the proposed approach applicable for regression domains having more than two dimensions. Using three spatial datasets and three higher dimensional datasets, we investigate the numerical performance of the approach and compare it to several state-of-the-art approaches.
\end{abstract}


\section{Introduction}
Gaussian process (GP) regression is a popular Bayesian nonparametric approach for nonlinear regression  \citep{rasmussen2006gaussian}. A GP prior is assumed for the unknown regression function, and the posterior estimate of the function is from this prior, combined with noisy (or noiseless, for deterministic simulation response surfaces) response observations. The posterior estimate can be easily derived in a simple closed form using the properties induced by the GP prior, and the estimator has several desirable properties, e.g., it is the best linear unbiased estimator under the assumed model and offers convenient quantification of the prediction error uncertainty. Its conceptual simplicity and attractive properties are major reasons for its popularity. On the other hand, the computational expense for evaluating the closed form solution is proportional to $N^3$, where $N$ denotes the number of observations, which can be prohibitively expensive for large $N$. Broadly speaking, this paper concerns fast computation of the GP regression estimate for large $N$. 

The major computational bottleneck for GP regression is the inversion of a $N \times N$ sample covariance matrix, which is also often poorly numerically conditioned. Different approaches for representing or approximating the sample covariance matrix with a more efficiently invertible form have been proposed. The approaches can be roughly categorized as sparse approximations, low-rank approximations, or local approximations. Sparse methods represent the sample covariance with a sparse version, e.g. by applying a covariance tapering technique \citep{furrer:06, kaufman2008covariance}, using a compactly supported covariance function \citep{gneiting2002compactly}, or using a Gaussian Markov approximation of a GP \citep{lindgren2011}. The inversion of a  sparse positive definite matrix is less computationally expensive than the inversion of a non-sparse matrix of the same size, because its Cholesky decomposition is sparse and can be achieved more quickly.

Low-rank approximations of the sample covariance matrix can be performed in multiple ways. The most popular approach for the low-rank approximation introduces latent variables and assume a certain independence conditioned on the latent variables \citep{seeger:03, snelson:06}, so that the resulting sample covariance matrix has reduced rank. The (pseudo)inversion of a $N \times N$ matrix of rank $M$ can be computed with reduced $O(NM^2)$ expense. \citet{titsias2009variational} introduced a variational formulation to infer the latent variables along with covariance parameters, and a variant of the idea was proposed using the stochastic variational inference technique \citep{Hensman13}. The latent variable model approaches are exploited to develop parallel computing algorithms for GP regression \citep{Chen:2013}. Another way for low rank approximation is to approximate the sample covariance matrix with a product of a block diagonal matrix and multiple blocked low-rank matrices \citep{ambikasaran2016fast}. 

Local approximation approaches partition the input domain into a set of local regions and assume an independent GP regression model within each region \citep{das2010block}. The resulting sample covariance matrix is a block diagonal matrix of local sample covariance matrices, and inverting the block diagonal matrix is much cheaper computationally. Such local approximation approaches have many advantages. By their local nature, they adapt better to local and nonstationary data features, and independent local approximation models can be computed in parallel to reduce total computation time. Their major weakness is that two local models for two neighboring local regions produce different predictions at the boundary between the regions, creating discontinuity of the regression function over the boundary. This boundary discontinuity is not just an aesthetic problem, as it was empirically shown that greater discontinuity implies greater degradation in prediction accuracy, particularly around the boundaries of the local regions \citep{park:16}. This discontinuity issue has been addressed in different ways. Perhaps the most popular approach is to smooth out some of the discontinuity by using some weighted average across the local models or across multiple sets of local models via a Dirichlet mixture \citep{rasmussen:02}, a treed mixture \citep{gramacy2008bayesian}, Bayesian model averaging \citep{tresp2000bayesian, chen2009bagging, deisenroth2015}, or locally weighted projections \citep{nguyen2009local}. Other, related approaches use additive covariance functions consisting of a global covariance and a local covariance \citep{snelson:07, vanhatalo08}, construct a local model for each testing point \citep{gramacy2014local}, or use a local partition but constrain the local models for continuity \citep{park:11, park:16}. 

In this work we use a partitioned input domain like \citet{park:11} and \citet{park:16}, but we introduce a different form of continuity constraints that are more easily and more naturally integrated into the GP modeling framework. Both \citet{park:11} and \citet{park:16} basically reformulated local GP regression as an optimization problem, and the local GP models for neighboring regions were constrained to have the same predictive means on the boundaries of the local regions by adding some linear constraints to the optimization problems that infer the local GP models. \citet{park:11} used a constrained quadratic optimization that constrains the predictive means for a finite number of boundary locations, and \citet{park:16} introduced a functional optimization formulation to enforce the same constraints for all boundary locations. The optimization-based  formulations make it infeasible to derive the marginal likelihood and the predictive variances in closed forms, which were roughly approximated. In contrast, this paper presents a simple and natural way to enforce continuity. We consider a set of GPs that are defined as the differences between the responses for the local GPs in neighboring regions. Continuity implies that these differenced GPs are identically zero along the boundary between neighboring regions. Hence, we impose continuity constraints by treating the values of the differenced GPs at a specified set of boundary points as all having been ``observed to be zero", and we refer to these zero-valued differences as pseudo-observations. We can then conveniently incorporate continuity constraints by simply augmenting the actual set of response observations with the set of pseudo-observations, and then using standard GP modeling to calculate the posterior predictive distribution given the augmented set of observations. We note that \textit{observing} the differenced GPs to be zero at a set of boundary points is essentially equivalent to \textit{assuming} continuity at these points without imposing any further assumptions on the nature of the GPs. 

The new modeling approach creates several major benefits over the previous domain partitioning approaches. The new modeling is simpler than the previous approaches, so the marginal likelihood function can be explicitly derived for tuning hyperparameters, which was not possible for the previous approaches. In the previous approaches, the values of the predictive means on the boundaries of local regions must be explicitly specified, which involves solving a large linear system of equations. Unlike the previous approaches, observing the pseudo-observations of the differenced GPs to be zero does not require specifying the actual values of the predictive means and variances on the boundaries. Furthermore, the proposed approach enforces that the local models for neighboring regions produce the same values for both the predictive means and variances at the boundary points between the local regions, while both of the previous approaches are only able to enforce the same predictive means but not the same predictive variances. Last, the previous approaches are only applicable for one- or two-dimensional problems, while our new approach is applicable for higher dimensional regression problems. We view our approach as “patching” together a collection of local GP regression models using the boundary points as “stitches”, and, hence, we refer to it as \textit{patchwork kriging}. 

The remainder of the paper is organized as follows. Section \ref{sec:gpr} briefly reviews the general GP regression problem and notational convention. Section~\ref{sec:method} presents the core methodology of the patchwork kriging approach, including the prior model assumptions, the pseudo-observation definition, the resulting posterior predictive mean and variance equations, and the detailed computation steps along with choice of tuning parameters. Section~\ref{sec:example} shows how the patchwork kriging performs with a toy example for illustrative purpose. Section \ref{sec:simstudy} investigates the numerical performance of the proposed method for different simulated cases and compares it with the exact GP regression (i.e., the GP regression without partitions, using the entire dataset to predict each point) and a global GP approximation method. Section \ref{sec:validation} presents the numerical performance of the proposed approach for five real datasets and compares it with \citet{park:16} and other state-of-the-art methods. Finally, Section \ref{sec:conc} concludes the paper with a discussion.

\section{Gaussian Process Regression} \label{sec:gpr}
Consider the general regression problem of estimating an unknown predictive function $f$ that relates a $d$ dimensional predictor $x \in \mathbb{R}^d$ to a real response $y$, using noisy observations $\dataset = \{(x_i,y_i), i=1,\ldots, N\}$,
\begin{equation*}
y_i = \mu+ f(x_i) + \epsilon_i, \qquad i =1, \dots, N,
\end{equation*}
where $\epsilon_i \sim \mathcal{N}(0, \sigma^2)$ is white noise, independent of $f(x_i)$. We assume that $\mu = 0$. Otherwise, one can normalize $y_i$ by subtracting the sample mean of the $y_i$'s from $y_i$. Notice that we do not use bold font for the multivariate predictor $x_i$ and reserve bold font for the collection of observed predictor locations, $\X{x} = [x_1,x_2,\dots,x_N]^T$.

In a GP regression for this problem, one assumes that $f$ is a realization of a zero-mean Gaussian process having covariance function $c(\cdot, \cdot)$ and then uses the observations $\dataset$ to obtain the posterior predictive distribution of $f$ at an arbitrary $x_*$, denoted by $f_* = f(x_*)$. Denote $\V{y} = [y_1,y_2,\dots,y_N]^T$. The joint distribution of $(f_*, \V{y})$ is
\begin{equation*}
P(f_*, \V{y}) =
\mathcal{N}\left( \V{0}, \left[ \begin{array} {c c} c_{**}
& \V{c}_{\X{x}*}^T \\ \V{c}_{\X{x}*}
& \sigma^2\M{I} + \M{C}_{\X{xx}} \end{array} \right] \right),
\end{equation*}
where $c_{**} = c(x_*, x_*)$, $\V{c}_{\X{x}*} = (c(x_1,x_*), \dots, c(x_N, x_*))^T$ and $\M{C}_{\X{xx}}$ is an $N \times N$ matrix with $(i,j)^{th}$ entry $c(x_i, x_j)$. The subscripts on $c_{**}, \V{c}_{\X{x}*}$, and $\M{C}_{\X{xx}}$ indicate the two sets of locations between which the covariance is computed, and we have abbreviated the subscript $x_*$ as $*$. Applying the Gaussian conditioning formula to the joint distribution gives the predictive distribution of $f_*$ given $\V{y}$ \citep{rasmussen2006gaussian},
\begin{equation}\label{eq:pred-dist}
P(f_* | \V{y})
= \mathcal{N}(
\V{c}_{\X{x}*}^T (\sigma^2\M{I} + \M{C}_{\X{xx}})^{-1} \V{y},
c_{**} - \V{c}_{\X{x}*}^T (\sigma^2\M{I} + \M{C}_{\X{xx}})^{-1}
\V{c}_{\X{x}*}).
\end{equation}
The predictive mean
$\V{c}_{\X{x}*}^T (\sigma^2\M{I} + \M{C}_{\X{xx}})^{-1} \V{y}$
is taken to be the point prediction of $f(x)$ at location $x_*$,
and its uncertainty is measured by the predictive variance
$c_{**} - \V{c}_{\X{x}*}^T (\sigma^2\M{I} + \M{C}_{\X{xx}})^{-1}
\V{c}_{\X{x}*}$. Efficient calculation of the predictive mean and variance for large datasets has been the focus of much research.

\section{Patchwork Kriging} \label{sec:method}
As mentioned in the introduction, for efficient computation we replace the GP regression by a set of local GP models on some partition of the input domain, in a manner that enforces a level of continuity in the local GP model responses over the boundaries separating their respective regions. Section \ref{sec:inf} conveys the main idea of the proposed approach, 

\subsection{Inference with Boundary Continuity Constraints} \label{sec:inf}
To specify the idea more precisely, consider a spatial partition of the input domain of $f(x)$ into $K$ local regions $\{\Omega_k:  k=1,2,...,K \}$, and define $f_k(x)$ as the local GP approximation of $f(x)$ at $x \in \overline{\Omega}_k$, where $\overline{\Omega}_k$ is the closure of $\Omega_k$. Temporarily ignoring the continuity requirements, the local models are assumed to follow independent GP priors:
\begin{assumption}
Each $f_k(x)$ follows a GP prior distribution with zero mean and covariance function $c_k(\cdot, \cdot)$, and the $f_k(x)$'s are mutually independent \textit{a priori} (prior to enforcing the continuity conditions). The choice of the local covariance function(s) can differ depending on the specifics of the problem. If $f(x)$ is expected to be a stationary process, then one could use the same $c_k(\cdot, \cdot) = c(\cdot, \cdot)$ for all $k$. In this case, the purpose of this local GP approximation would be to approximate $f(x)$  computationally efficiently. On the other hand, if one expects non-stationary behavior of the data, then different covariance functions should be used for each region. 
\end{assumption}

It is important to note that the independence of the GPs in Assumption 1 is prior to enforcing the continuity conditions via the pseudo-observations, as described below. After enforcing the continuity conditions, the GPs will no longer be independent \textit{a priori}, since the assumed continuity at the boundaries imposes a very strong prior dependence of the surfaces.  Since the pseudo-observations should also be viewed as additional prior information, the independence condition in Assumption 1 might be more appropriately viewed as a \textit{hyperprior} condition.  In fact, we view the incorporation of the boundary pseudo-observations as an extremely tractable and straightforward way of imposing some reasonable form of dependency of the $f_k(x)$ across regions (which is the ultimate goal), while still allowing us to begin with an independent GP \textit{hyperprior} (which results in the tractability of the analyses).

Now partition the training set $\mathcal{D}$ into $\mathcal{D}_k = \{(x_i,y_i): x_i \in \Omega_k\}$ $(k = 1, 2, . . ., K)$, and denote $\X{x}_k = \{ x_i: x_i \in \Omega_k \}$ and $\V{y}_k = \{ y_i: x_i \in \Omega_k\}$. By the independence part of Assumption 1, the predictive distribution of $f_k(x)$ given $\mathcal{D}$ is equivalent to the predictive distribution of $f_k(x)$ given $\mathcal{D}_k$, which gives the standard local GP solution with no continuity requirements. The primary problem with this formulation is that the predictive distributions of $f_k(x)$ and $f_l(x)$ are not equal on the boundary of their neighboring regions $\Omega_k$ and $\Omega_l$.
\begin{figure}[h]
    \centering
            \includegraphics[width=0.7\textwidth]{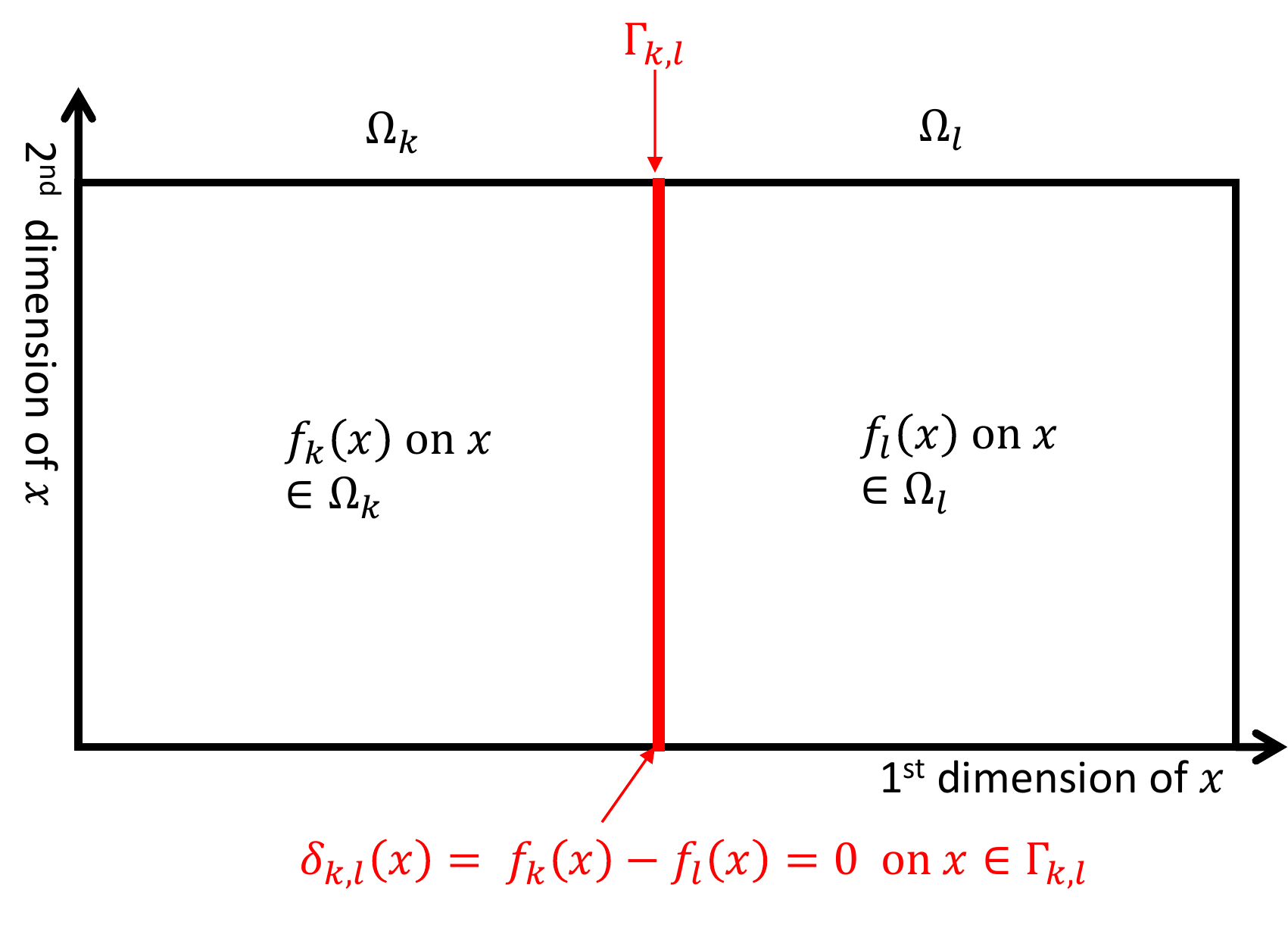}
    \caption{Illustration of the notation and concepts for defining the local models $f_k(x)$ and $f_l(x)$. $\Omega_k$ and $\Omega_l$ represent two local regions resulting from some appropriate spatial partition (discussed later) of the regression input domain. The (posterior distributions for the) GP functions $f_k(x)$ and $f_l(x)$ represent the local approximations of the regression function on $\Omega_k$ and $\Omega_l$, respectively. The subset $\Gamma_{k,l}$ represents the interfacial boundary between the two regions $\Omega_k$ and $\Omega_l$, and $\delta_{k,l}(x)$ is defined as the difference between $f_k(x)$ and $f_l(x)$, which is identically zero on $\Gamma_{k,l}$ by the continuity assumption.}
    \label{fig:notations}
\end{figure}

Our objective is to improve the local kriging prediction by enforcing $f_k(x)=f_l(x)$ on their shared boundary. The key idea is illustrated in Figure \ref{fig:notations} and described as follows. For two neighboring regions $\Omega_k$ and $\Omega_l$, let $\Gamma_{k,l} = \overline{\Omega}_k \cap \overline{\Omega}_l$ denote their shared boundary. For each pair of neighboring regions $\Omega_k$ and $\Omega_l$, we define the auxiliary process $\delta_{k,l}(x)$ to be the difference between the two local GP models, 
\begin{equation}
\delta_{k,l}(x) = f_k(x) - f_l(x) \mbox{ for } x \in \Gamma_{k,l},
\end{equation} 
and it is only defined for $k < l$ to avoid any duplicated definition of the auxiliary process. By the definition and under Assumption 1, $\delta_{k,l}(x)$ is a Gaussian process with zero mean and covariance function $c_k(\cdot, \cdot) + c_l(\cdot, \cdot)$, and its covariance with $f_j(x)$ is
\begin{equation*}
\begin{split}
\CV(\delta_{k,l}(x_1), f_j(x_2)) & = \CV(f_k(x_1) - f_l(x_1), f_j(x_2)) \\
                                 & = \CV(f_k(x_1), f_j(x_2)) - \CV(f_l(x_1), f_j(x_2)).
\end{split}
\end{equation*}
Since $\CV(f_k(x_1), f_l(x_2)) = c_k(x_1, x_2)$ for $k=l$ and zero otherwise under Assumption 1, 
\begin{equation} \label{eq:fd_rel}
\CV(\delta_{k,l}(x_1), f_j(x_2)) =\begin{cases}
c_k(x_1, x_2) &  \mbox{ if } k = j\\
-c_l(x_1, x_2)& \mbox{ if } l = j \\
0& \mbox{ otherwise}.
\end{cases}
\end{equation}
Likewise, $\delta_{k,l}(x)$ and $\delta_{u,v}(x)$ are correlated with covariance 
\begin{equation} \label{eq:dd_rel}
\CV(\delta_{k,l}(x_1), \delta_{u,v}(x_2)) = \begin{cases}
    c_k(x_1, x_2)& \text{if } k = u, l\neq v \\
    c_l(x_1, x_2)& \text{if } l = v, k \neq u \\
    -c_k(x_1, x_2)& \text{if } k = v, l \neq u \\
    -c_l(x_1, x_2)& \mbox{if } l = u, k \neq v\\
    c_k(x_1, x_2) +c_l(x_1, x_2) & \text{if } k = u, l = v \\
    0 & \text{otherwise.}
\end{cases}
\end{equation}

Boundary continuity between $f_k(x)$ and $f_l(x)$ can be achieved by enforcing the condition $\delta_{k,l}(x) = 0$ at $\Gamma_{j,k}$. We reiterate that $f_k(x)$ and $f_l(x)$ are no longer independent after conditioning on the additional information $\delta_{k,l}(x) = 0$. In fact, they are strongly dependent, as they must be in order to achieve continuity \textit{a priori}. Hence, the independence condition in Assumption 1 is really a \textit{hyperprior} independence. 

Deriving the exact prior distribution of the surface conditioned on $\delta_{k,l}(x) = 0$ everywhere on the boundaries appears to be computationally intractable, because there are uncountably infinitely many $x$'s in $\Gamma_{j,k}$. Instead, we propose a much simpler continuity correction that begins with the Assumption 1 prior (including the independence \textit{hyperprior}) and augments the observed data $\dataset$ with the pseudo observations $\delta_{k,l}(x) = 0$ for a finite number of input locations $x \in \Gamma_{k,l}$. As the number of boundary pseudo-observations grows, we can better approximate the theoretical ideal condition that $\delta_{k,l}(x) = 0$ everywhere on the boundary. The choice of the boundary input locations will be discussed later. 

Notice that observing ``$\delta_{k,l}(x) = 0$" is equivalent to observing that $f_k(x) = f_l(x)$ without observing the actual values of $f_k(x)$ and $f_l(x)$. Thus, if we augment $\dataset$ to include these pseudo observations when calculating the posterior predictive distributions of $f_k(x)$ and $f_l(x)$, it will force the posterior distributions of $f_k(x)$ and $f_l(x)$ to be the same at each boundary input location $x$, because observing $\delta_{k,l}(x) = 0$ means that we have observed $f_k(x)$ and $f_l(x)$ to be the same (see \eqref{eq:cont1} and \eqref{eq:cont2}, for a formal proof). This implies that their posterior means (which are the GP regression predictive functions) and their posterior variances (which quantify the uncertainty in the predictions) will both be equal. 

Suppose that we place $B$ pseudo observations on each $\Gamma_{k,l}$. Let $\V{x}_{k,l}$ denote the set of $B$ input boundary locations chosen in $\Gamma_{k,l}$, let $\V{\delta}_{k,l}$ denote a collection of the noiseless observations of $\delta_{k,l}(x)$ at the selected boundary locations, and let $\V{\delta}$ denote the collection of all $\V{\delta}_{k,l}$'s in the following order,
\begin{equation*}
\V{\delta}^T = (\V{\delta}_{1,1}^T, \V{\delta}_{1,2}^T, \ldots, \V{\delta}_{1,K}^T,\V{\delta}_{2,1}^T, \ldots, \V{\delta}_{2,K}^T,\ldots, \V{\delta}_{KK}^T).
\end{equation*}
 Note that the observed pseudo value of $\V{\delta}$ will be a vector of zeros, but its prior distribution (prior to observing the pseudo values or any response observations) is represented by the above covariance expressions. Additionally, let $f_{*}^{(k)} = f_k(x_*)$ denote the value of the response $f_k(x)$ at any $x_* \in \Omega_k$, and let $\V{y}^T = (\V{y}_1^T, \V{y}_2^T, \ldots, \V{y}_K^T)$. The prior joint distribution of $f_{*}^{(k)}$, $\V{y}$ and $\V{\delta}$ is 
\begin{equation} \label{eq:joint}
\left[\begin{array}{c}
f_{*}^{(k)} \\ \V{y} \\ \V{\delta}
\end{array}\right]
\sim \mathcal{N}\left(     
\left[\begin{array}{c}
0 \\ \V{0} \\ \V{0} 
\end{array}\right],
\left[\begin{array}{c c c}
c_{**}       & \V{c}_{*\dataset}^{(k)} &  \V{c}_{*, \delta}^{(k)}\\
\V{c}_{\dataset*}^{(k)}   & \V{C}_{\dataset\dataset}      & \V{C}_{\dataset,\delta}  \\
\V{c}_{\delta,*}^{(k)} & \V{C}_{\delta,\dataset}  & \V{C}_{\delta,\delta} \\
\end{array}\right]
\right),
\end{equation}
where the expressions of the covariance blocks are given by $c_{**} = \CV(f^{(k)}_*, f^{(k)}_*)$,
\begin{equation*}
\begin{split}
& \V{c}_{*\dataset}^{(k)} = (\CV(f^{(k)}_*, \V{y}_1), \CV(f^{(k)}_*, \V{y}_2), \ldots, \CV(f^{(k)}_*, \V{y}_K)),\\
& \V{c}_{*\delta}^{(k)} = (\CV(f_{*}^{(k)}, \V{\delta}_{1,1}), \CV(f_{*}^{(k)}, \V{\delta}_{1,2}), \ldots, \CV(f_{*}^{(k)}, \V{\delta}_{K,K})),
\end{split}
\end{equation*}
\begin{equation*}
\V{c}_{\dataset\dataset} = \left[\begin{array}{c c c c}
\CV(\V{y}_1, \V{y}_1) & \CV(\V{y}_1, \V{y}_2) & \ldots & \CV(\V{y}_1, \V{y}_K) \\
\CV(\V{y}_2, \V{y}_1) & \CV(\V{y}_2, \V{y}_2) & \ldots & \CV(\V{y}_2, \V{y}_K) \\
\vdots& \vdots & \ddots & \vdots \\
\CV(\V{y}_K, \V{y}_1) & \CV(\V{y}_K, \V{y}_2) & \ldots & \CV(\V{y}_K, \V{y}_K)
\end{array}\right],
\end{equation*}
\begin{equation*}
\hspace{50pt} \M{C}_{\dataset,\delta} = \left[\begin{array}{c c c c}
\CV(\V{y}_1, \V{\delta}_{1,1}) & \CV(\V{y}_1, \V{\delta}_{1,2}) & \ldots &  \CV(\V{y}_1, \V{\delta}_{K,K}) \\
\CV(\V{y}_2, \V{\delta}_{1,1}) & \CV(\V{y}_2, \V{\delta}_{1,1}) & \ldots &  \CV(\V{y}_2, \V{\delta}_{K,K}) \\
\vdots& \vdots & \ddots & \vdots \\
\CV(\V{y}_K, \V{\delta}_{1,1}) & \CV(\V{y}_K, \V{\delta}_{1,2}) & \ldots &  \CV(\V{y}_K, \V{\delta}_{K,K})
\end{array}\right], \mbox{ and }
\end{equation*}
\begin{equation*}
\hspace{50pt} \M{C}_{\delta,\delta} = \left[\begin{array}{c c c c}
\CV(\V{\delta}_{1,1}, \V{\delta}_{1,1}) & \CV(\V{\delta}_{1,1}, \V{\delta}_{1,2}) & \ldots &  \CV(\V{\delta}_{1,1}, \V{\delta}_{K,K}) \\
\CV(\V{\delta}_{1,2}, \V{\delta}_{1,1}) & \CV(\V{\delta}_{1,2}, \V{\delta}_{1,1}) & \ldots &  \CV(\V{\delta}_{1,2}, \V{\delta}_{K,K}) \\
\vdots& \vdots & \ddots & \vdots \\
\CV(\V{\delta}_{K,K}, \V{\delta}_{1,1}) & \CV(\V{\delta}_{K,K}, \V{\delta}_{1,2}) & \ldots &  \CV(\V{\delta}_{K,K}, \V{\delta}_{K,K})
\end{array}\right].
\end{equation*}
Note that joint covariance matrix is very sparse, because of the many zero values in \eqref{eq:fd_rel} and \eqref{eq:dd_rel}.

From the standard GP modeling results applied to the augmented data, the posterior predictive distribution of $f_{*}^{(k)}$ given $\V{y}$ and the pseudo observations $\V{\delta} = \V{0}$ is Gaussian with mean
\begin{equation} \label{eq:posmean}
\begin{split} 
\EX[f_{*}^{(k)}|\V{y}, \V{\delta}=\V{0}] = (\V{c}_{*\dataset}^{(k)}- \V{c}_{*\delta}^{(k)}\V{C}_{\delta,\delta}^{-1}\M{C}_{\dataset\delta}^T)(\M{C}_{\dataset\dataset}-\M{C}_{\dataset\delta}\V{C}_{\delta,\delta}^{-1}\M{C}_{\dataset\delta}^T)^{-1}\M{y}.
\end{split}
\end{equation}
and variance 
\begin{equation} \label{eq:posvar}
\begin{split}
\VA[f_{*}^{(k)}|\V{y}, \V{\delta}] 
& = c_{**}- \V{c}_{*, \delta}^{(k)} \V{C}_{\delta,\delta}^{-1}\V{c}_{\delta*}^{(k)} \\
  & \quad -  (\V{c}_{*\dataset}^{(k)} - \V{c}_{*\delta}^{(k)}\V{C}_{\delta,\delta}^{-1}\M{C}_{\dataset\delta}^T)(\M{C}_{\dataset\dataset}-\M{C}_{\dataset\delta}\V{C}_{\delta,\delta}^{-1}\M{C}_{\dataset\delta}^T)^{-1}(\V{c}_{\dataset*}^{(k)} - \V{C}_{\dataset\delta}\V{C}_{\delta,\delta}^{-1}\M{c}_{\delta*}^{(k)}).
\end{split}
\end{equation}
The derivation of the predictive mean and variance can be found in Appendix A.

One implication of the continuity imposed by including the pseudo observations $\V{\delta} = \V{0}$ is that the posterior predictive means and variances of $f_{*}^{(k)}$ and $f_{*}^{(l)}$ for two neighboring regions $\Omega_k$ and $\Omega_l$ are equal at the specified input boundary locations $\X{x}_{k,l}$; see Appendix B for details. The continuity imposed certainly does not guarantee that the posterior means and variances of $f_{*}^{(k)}$ and $f_{*}^{(l)}$ are equal for every $x^* \in \Gamma_{k,l}$, including those not in the locations of pseudo observations $\X{x}_{k,l}$. Our numerical experiments in Section \ref{sec:simstudy} demonstrate that as we place more pseudo inputs, the posterior means and variances of $f_{*}^{(k)}$ and $f_{*}^{(l)}$ converge to each other.

From the preceding, our proposed approach enforces that the two local GP models for two neighboring local regions have the same posterior predictive means and variances (and they satisfy an even stronger condition, that the responses themselves are identical) at the chosen set of boundary points corresponding to the pseudo observations. We view this as patching together the independent local models in a nearly continuous way. The chosen sets of boundary points serve as the stitches when patching together the pieces, and the more boundary points are chosen, the closer the models are to being continuous over the entire boundary. In light of this, we refer to the approach as \textit{patchwork kriging}. 

\subsection{Hyperparameter Learning and Prediction} \label{sec:algorithm}
The hyperparameters of the covariance function(s) $c_k(\cdot, \cdot)$ determine the correlation among the values of $f(x)$, which has significant effect on the accuracy of a Gaussian process regression. We jointly estimate all correlation parameters (multiple sets of parameters if different $c_k(\cdot, \cdot)$ are assumed for each region) using maximum likelihood estimation (MLE) by minimizing the negative log marginal likelihood, 
\begin{equation} \label{eq:mle}
\begin{split}
NL(\theta)  = & - \log p(\V{y}, \V{\delta} = \V{0}|\theta) \\
            = & \frac{N}{2}\log (2\pi) + \frac{1}{2} \log \left|\begin{array}{c c}
            \V{C}_{\dataset\dataset}      & \V{C}_{\dataset,\delta}  \\
            \V{C}_{\delta,\dataset}  & \V{C}_{\delta,\delta} \\
            \end{array}\right| + \frac{1}{2} [\V{y}^T \V{0}^T] \left[\begin{array}{c c}
            \V{C}_{\dataset\dataset}      & \V{C}_{\dataset,\delta}  \\
            \V{C}_{\delta,\dataset}  & \V{C}_{\delta,\delta} \\
            \end{array}\right]^{-1} \left[\begin{array}{c}
            \V{y}  \\
            \V{0} \\
            \end{array}\right]
\end{split}
\end{equation}
Note that we have augmented the data to included the pseudo observations $\V{\delta} = \V{0}$ in the likelihood expression, which results in a better behaved likelihood by imposing some continuity across regions. This essentially allows data to be shared across regions when estimating the covariance parameters. Using the properties of a determinant for a partitioned matrix, the determinant part in the marginal likelihood becomes
\begin{equation*}
\left|\begin{array}{c c}
\V{C}_{\dataset\dataset}      & \V{C}_{\dataset,\delta}  \\
\V{C}_{\delta,\dataset}  & \V{C}_{\delta,\delta} \\
\end{array}\right| = |\V{C}_{\dataset\dataset}| |\V{C}_{\delta,\delta} - \V{C}_{\delta,\dataset}\V{C}_{\dataset\dataset}^{-1} \V{C}_{\dataset,\delta}|,
\end{equation*}
which can be utilized to compute the log determinant term in $NL(\theta)$ as follows,
\begin{equation*}
\log \left|\begin{array}{c c}
            \V{C}_{\dataset\dataset}      & \V{C}_{\dataset,\delta}  \\
            \V{C}_{\delta,\dataset}  & \V{C}_{\delta,\delta} \\
            \end{array}\right| = 
\log \left|\V{C}_{\dataset\dataset}\right| + \log \left|\V{C}_{\delta,\delta} - \V{C}_{\delta,\dataset}\V{C}_{\dataset\dataset}^{-1} \V{C}_{\dataset,\delta}\right|.
\end{equation*}
Note that the log determinant of the block diagonal matrix $\V{C}_{\dataset\dataset}$ is equal to the sum of the log determinants of its diagonal blocks, and $\V{C}_{\delta,\delta} - \V{C}_{\delta,\dataset}\V{C}_{\dataset\dataset}^{-1} \V{C}_{\dataset,\delta}$ is very sparse, so the cholesky decomposition of the sparse matrix can be taken to evaluate their determinants; we will detail the sparsity discussion in the next section. 
Evaluating the quadratic term of the negative log marginal likelihood function involves the inversion of $(\M{C}_{\dataset\dataset}-\M{C}_{\dataset\delta}\V{C}_{\delta,\delta}^{-1}\M{C}_{\dataset\delta}^T)$. The inversion can be effectively evaluated using 
\begin{equation} \label{eq:Cinv}
(\M{C}_{\dataset\dataset}-\M{C}_{\dataset\delta}\V{C}_{\delta,\delta}^{-1}\M{C}_{\dataset\delta}^T)^{-1} =  \M{C}_{\dataset\dataset}^{-1} + \M{C}_{\dataset\dataset}^{-1} \M{C}_{\dataset\delta} ( \V{C}_{\delta,\delta} -  \M{C}_{\dataset\delta}^T\M{C}_{\dataset\dataset}^{-1}\M{C}_{\dataset\delta} )^{-1} \M{C}_{\dataset\delta}^T \M{C}_{\dataset\dataset}^{-1}.
\end{equation}

After the hyperparameters were chosen by the MLE criterion, evaluating the predictive mean and variance for the patchwork kriging model can be performed as follows. Let $\M{Q}$ denote the inversion result of \eqref{eq:Cinv}, and let $\M{L}$ denote the cholesky decomposition of $\V{C}_{\delta,\delta}$ such that $\V{C}_{\delta,\delta} = \M{L} \M{L}^T$.  After the pre-computation of the two matrices and $\V{v} = \M{L}^{-1} \M{C}_{\dataset \delta}^T$, the predictive mean \eqref{eq:posmean} and the predictive variance \eqref{eq:posvar} can be evaluated for each $x_* \in \Omega_k$,
\begin{equation}
\begin{split}
& \EX[f_{*}^{(k)}|\V{y}, \V{\delta}=\V{0}] = (\V{c}_{*\dataset}^{(k)} - \V{w}_*^T \V{v}) \M{Q} \V{y} \\
& \VA[f_{*}^{(k)}|\V{y}, \V{\delta}=\V{0}] = c_{**} - \V{w}_*^T \V{w}_* - (\V{c}_{*\dataset}^{(k)} - \V{w}_*^T \V{v}) \M{Q} (\V{c}_{*\dataset}^{(k)} - \V{w}_*^T \V{v})^T,
\end{split}
\end{equation}
where $\V{w}_* = \M{L}^{-1} (\V{c}_{*\delta}^{(k)})^T$. The computation steps of patchwork kriging were described in Algorithm \ref{alg:comp}. 
\begin{algorithm}[t]
    \caption{Computation Steps for Patchwork Kriging} 
    \label{alg:comp}
  \begin{algorithmic}[1]
    \REQUIRE \\
    Decomposition of domain $\{\Omega_k; k=1,\ldots, K\}$; see Section \ref{sec:partition} for a choice. \\
    Locations of pseudo data $\{\X{x}_{k,l}; k,l = 1,\ldots, K\}$; see Section \ref{sec:partition} for a choice. \\
    Hyperparameters of covariance function $c_k(\cdot, \cdot)$; use the MLE criterion \eqref{eq:mle} for a choice.
    \INPUT Data $\dataset$ and test location $x_*$
    \OUTPUT $\EX[f_{*}^{(k)}|\V{y}, \V{\delta}=\V{0}]$ and $\VA[f_{*}^{(k)}|\V{y}, \V{\delta}=\V{0}]$
    \STATE \textbf{Evaluate} $\mathbf{Q} = (\M{C}_{\dataset\dataset}-\M{C}_{\dataset\delta}\V{C}_{\delta,\delta}^{-1}\M{C}_{\dataset\delta}^T)^{-1}$ using \eqref{eq:Cinv}.
    \STATE \textbf{Take} the Cholesky Decomposition of $\V{C}_{\delta,\delta} = \M{L} \M{L}^T$.
    \STATE \textbf{Evaluate} $\V{v} = \M{L}^{-1} \M{C}_{\dataset \delta}^T$.
    \FOR{$x_* \in \Omega_k$}
         \STATE \textbf{Evaluate} $\V{w}_* = \M{L}^{-1} (\V{c}_{*\delta}^{(k)})^T$ .
         \STATE $\EX[f_{*}^{(k)}|\V{y}, \V{\delta}=\V{0}] = (\V{c}_{*\dataset}^{(k)} - \V{w}_*^T \V{v}) \M{Q} \V{y}$. 
         \STATE $\VA[f_{*}^{(k)}|\V{y}, \V{\delta}=\V{0}] = c_{**} - \V{w}_*^T \V{w}_* - (\V{c}_{*\dataset}^{(k)} - \V{w}_*^T \V{v}) \M{Q} (\V{c}_{*\dataset}^{(k)} - \V{w}_*^T \V{v})^T$.
    \ENDFOR
  \end{algorithmic}
\end{algorithm}

\subsection{Sparsity and Complexity Analysis} \label{sec:complexity}
The computational expense of patchwork kriging is much less than that of the global GP regression. The computational expense of the patchwork kriging model is dominated by evaluating the inversion \eqref{eq:Cinv}. The inversion comes in two parts. The first part is to invert $\M{C}_{\dataset\dataset}$. Note that $\M{C}_{\dataset\dataset}$ is a block diagonal matrix with the $k$th block size equal to the size of $\dataset_k$. If the size of each $\dataset_k$ is $M$, evaluating $ \M{C}_{\dataset\dataset}^{-1}$ requires only inverting $K$ matrices of size $M \times M$, and its expense is $O(KM^3)$. Given $ \M{C}_{\dataset\dataset}^{-1}$, evaluating $ \M{C}_{\dataset\dataset}^{-1} \M{C}_{\dataset\delta}$ adds $O(KBM^2)$ to the computational expense. 

The second part of the inversion \eqref{eq:Cinv} is to invert $\V{C}_{\delta,\delta} -  \M{C}_{\dataset\delta}^T\M{C}_{\dataset\dataset}^{-1}\M{C}_{\dataset\delta}$.  The matrix is very sparse, because $\CV(\V{\delta}_{k,l}, \V{\delta}_{u,v}) - \sum_{m=1}^K \CV(\V{\delta}_{k,l}, \V{y}_{m}) \CV(\V{y}_m, \V{y}_m)^{-1} \CV(\V{y}_{m}, \V{\delta}_{u,v})$ is a zero matrix unless the tuple $(k,l,u,v)$ satisfies the non-zero conditions listed in \eqref{eq:dd_rel}. The symmetric sparse matrix can be converted into a symmetric sparse banded matrix by the reverse Cuthill-McKee algorithm \citep{Chan1980}, and the computational complexity of the conversion algorithm is linearly proportional to the number of non-zero elements in the original sparse matrix. Let $d_f$ denote the number of neighboring local regions of each local region, and $B$ denote the number of pseudo observations placed per boundary. The number of non-zero elements in the sparse matrix is $O(d_fBK)$, so the time complexity of the reverse Cuthill-McKee algorithm is $O(d_fBK)$. The bandwidth of the resulting sparse matrix is linearly proportional to $d_f B$, and the size of the matrix is proportional to $d_fBK$. The complexity of taking the inverse of a symmetric banded matrix with size $r$ and bandwidth $p$ through Cholesky decomposition is $O(rp^2)$ \citep[pp. 154]{golub2012matrix}. Therefore, the complexity of inverting $\V{C}_{\delta,\delta} -  \M{C}_{\dataset\delta}^T\M{C}_{\dataset\dataset}^{-1}\M{C}_{\dataset\delta}$ is $O(d_f^3B^3K)$. Note that $d_f \propto d$ if data are more densely distributed over the entire input dimensions, and $d_f \propto d'$ if data are a $d'$-dimensional embedding in $d$ dimensional space.  The complexity becomes $O(d^3B^3K)$ for the worst case scenario.

Therefore, the total computational expense of the inversion \eqref{eq:Cinv} is $O(KM^3 + KBM^2 +  d_f^3B^3K)$. Typically, $B \ll M$, in which case the complexity is $O(KM^3  + d_f^3B^3K)$. Note that $M  \approx N/K$, where the approximation error is due to rounding $N/K$ to an integer value. Therefore, the complexity can be written as $O(N^3/K^2 + d_f^3B^3K)$. Given a fixed data size $N$, more splits of the regression domain will reduce the computation time due to the first term in the complexity, but too much increase would increase the second term, which will be shown later in Section \ref{sec:validation}. When data are dense in the input space, $d_f \propto d$, for which the complexity would increase in $d^3$. For $d < 10$, the effect is pretty ignorable unless $B$ is very large, but it becomes significant when $d > 10$. This computation issue related to data dimensions basically suggests to limit the practical use of this method to $d$ less than 100. We will later discuss more on this issue and how to choose $K$ and $B$ to balance off the computation and prediction accuracy in Section \ref{sec:KBstudy}.

In addition to the computational expense of the big inverse, additional incremental computations are needed per each test location $x_*$. The first part is to evaluate for the predictive mean and variances, 
\begin{equation} \label{eq:addpart}
(\V{c}_{*\dataset}^{(k)}- \V{c}_{*\delta}^{(k)}\V{C}_{\delta,\delta}^{-1}\M{C}_{\dataset\delta}^T).
\end{equation}
Note that the elements in $\V{c}_{*\dataset}^{(k)}$ are mostly zero except for the columns that correspond to $\dataset_k$ (size $M$), and similarly most elements of $\V{c}_{*\delta}^{(k)}$ are zero except for the columns that correspond to $\V{\delta}_{k,l}$'s (size $d_fB$). The cost of evaluating \eqref{eq:addpart} is $O(M + d_fB)$. Therefore, the cost of the predictive mean prediction per a test location is $O(M + d_fB)$, and the cost for the predictive variance is $O((M+d_fB)^2)$. When data are dense in the input dimensions, the costs become $O(M + dB)$ and $O((M+dB)^2)$.

\subsection{Tuning Parameter Selection} \label{sec:partition}
The performance of the proposed patchwork kriging method depends on the choice of tuning parameters, including the number of partitions ($K$) and the number ($B$) and locations ($\X{x}_{k,l}$) of the pseudo observations. This section presents guidelines for these choices. Choosing the locations of pseudo observations is related to the choice of domain partitioning. In this section, we discuss the choices of the locations and partitioning together. 

\begin{figure}[ht!]
        \centering
                \includegraphics[width=0.7\textwidth]{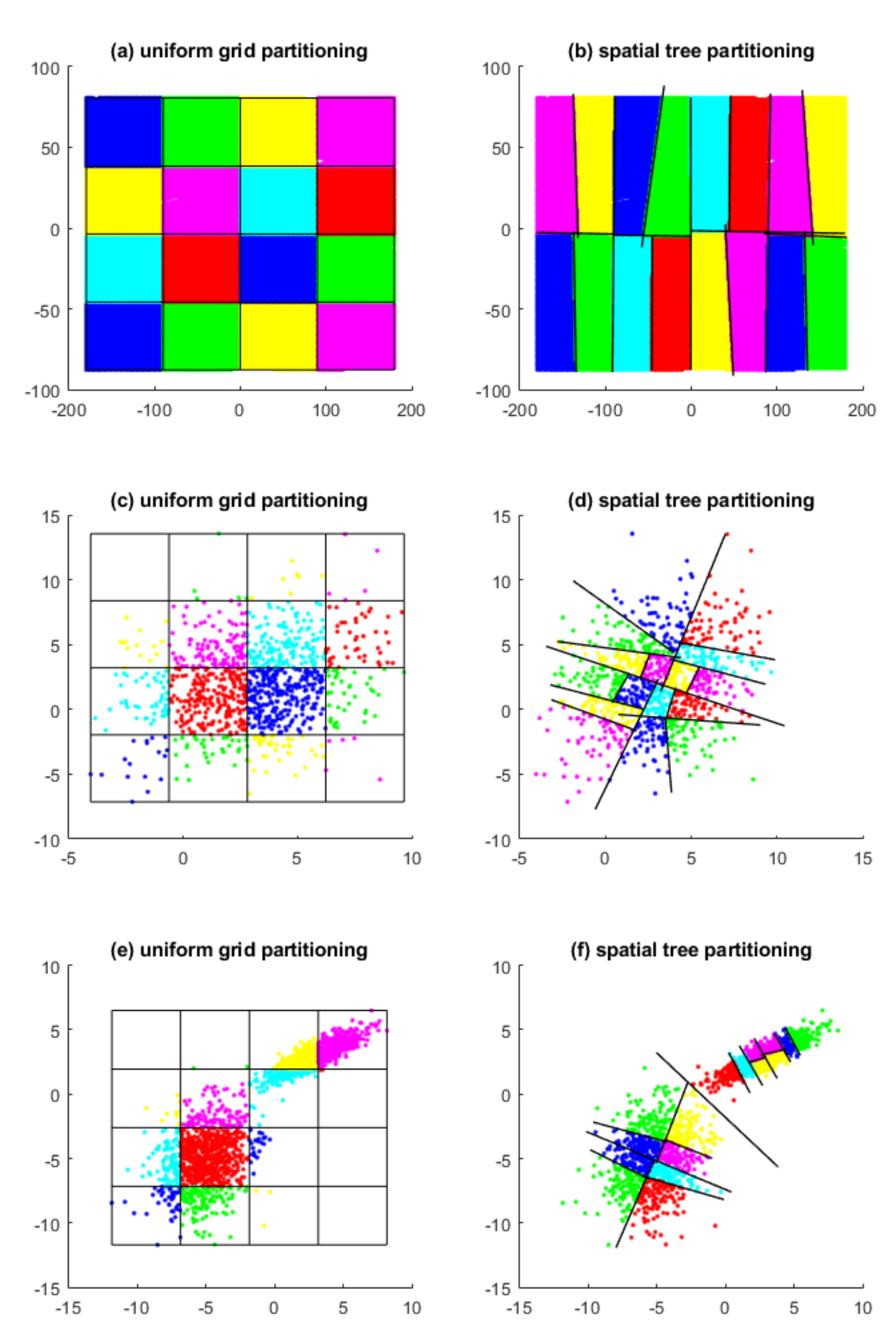}
        \caption{Comparison of two spatial partitioning schemes: a uniform grid (left panel) and a spatial tree (right panel). The spatial tree generates data partitioning of uniform sizes when data is unevenly distributed, but the uniform grid does not.}
        \label{fig:example_partition}
\end{figure}

There are many existing methods to partition a large set of data into smaller pieces. 
The simplest spatial partitioning is a uniform grid partitioning that divides a domain into uniform grids and splits data accordingly \citep{park:11, park:16}. This is simple and effective if the data are uniformly distributed over a low dimensional space. However, if the input dimension is high, it would either generate too many regions or it would produce many sparse regions that contain very few or no observations, and the latter also happens when the data are non-uniformly distributed; see examples in Figure \ref{fig:example_partition}-(c) and Figure \ref{fig:example_partition}-(e). \citet{shen2006fast} used a kd-tree for spatial partitioning of unevenly distributed data points in a high dimensional space. A kd-tree is a recursive partitioning scheme that recursively bisects the subspaces along one chosen data dimension at a time. Later, \citet{mcfee2011large} generalized it to the spatial tree. Starting with a level 0 space $\Omega^{(0)}_1$ equal to the entire regression domain, the spatial tree recursively bisects each of level $s$ spaces into two level $s+1$ spaces. Let $\Omega^{(s)}_j \in \mathbb{R}^d$ denote the $j$th region in the level $s$ space. It is bisected into two level $s+1$ spaces as
\begin{equation}
\Omega^{(s+1)}_{2j-1} = \{ \V{x} \in \Omega^{(s)}_j: \V{v}_{j,s}^T \V{x} \le \nu\} \mbox{ and } \Omega^{(s+1)}_{2j} = \{ \V{x} \in \Omega^{(s)}_j: \V{v}_{j,s}^T \V{x} > \nu\}.
\end{equation}
Each of $\Omega^{(s+1)}_{2j-1}$ and $\Omega^{(s+1)}_{2j}$ will be further partitioned in the next level using the same procedure. The choice of the linear projection vector $\V{v}_{j,s}$ depends on the distribution of the local data belonging to $\Omega^{(s)}_j$. For example, it can be the first principal component direction of the local data. The value for $\nu$ is chosen so that $\Omega^{(s+1)}_{2j-1}$ and $\Omega^{(s+1)}_{2j}$ have an equal number of observations. In this sense, the subregions at the same level are equally sized, which helps to level off the computation times of the local models. When the spatial tree is applied on data uniformly distributed over a rectangular domain, it produces a uniform grid partitioning; see examples in Figure \ref{fig:example_partition}-(b). The spatial tree is more effective than the grid partitioning when data is unevenly distributed; see examples in Figure \ref{fig:example_partition}-(d) and Figure \ref{fig:example_partition}-(f). 

In this work, we use a spatial tree with the principal component (PC) direction for $\V{v}_{j,s}$. Bisecting a space along the PC direction has effects of minimizing the area of the interfacial boundaries in between the two bisected regions, so the number of the pseudo observations necessary for connecting the two regions can be minimized. The maximum level of the recursive partitioning depends on the choice of $K$ via $s_{max} = \lfloor \log_2 K \rfloor$. The choices of $K$ and $B$ will be discussed in the next section. Given $B$, the pseudo observations $\X{x}_{k,l}$ are randomly generated from an uniform distribution over the intersection of the hyper-plane $\V{v}_{j,s}^T \V{x} = \nu$ and the level $s$ region $\Omega^{(s)}_j$.

\section{Illustrative Example} \label{sec:example}
\begin{figure}[h]
    \centering
            \includegraphics[width=\textwidth]{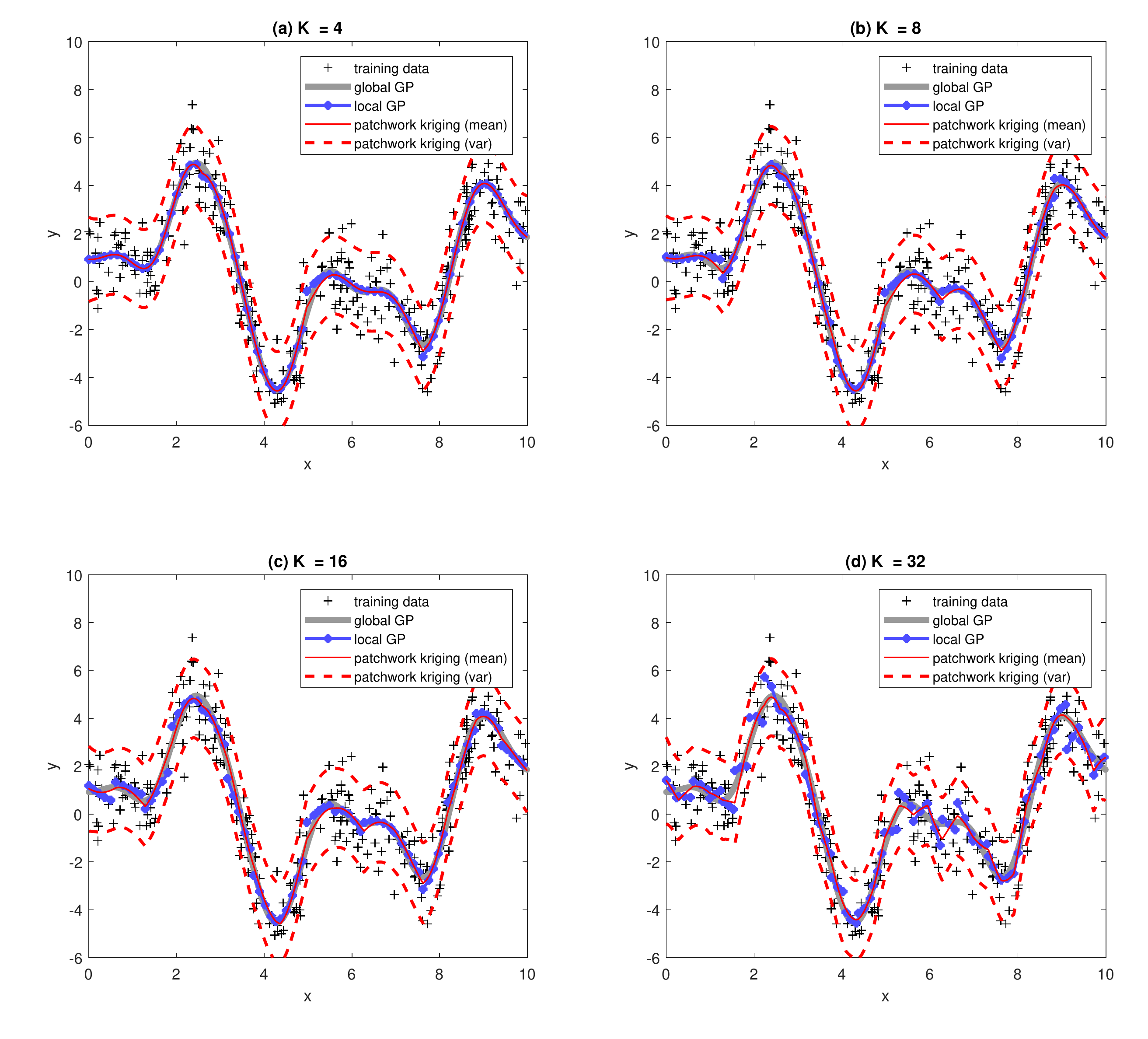}
    \caption{Example illustrating the patchwork kriging predictor, together with the global GP predictor and the regular local GP predictor with no continuity constraints. $K$ is the number of local regions.}
    \label{fig:1d_illustration}
\end{figure}

To illustrate how patchwork kriging changes the model predictions (relative to a set of independent GP models over each region, with no continuity conditions imposed), we designed the following simple simulation study; we will present more comprehensive simulation comparisons and analyses in Section 5. We generated a dataset of 6,000 noisy observations
\begin{equation*}
y_i = f(x_i) + \epsilon_i \quad \mbox{ for } i = 1,\ldots, 6000,
\end{equation*} 
 from a zero-mean Gaussian process with an exponential covariance function of $c(x_i, x_j) = 10\exp(-||x_i-x_j||_2)$, where $x_i \sim \mbox{Uniform}(0, 10)$ and $\epsilon_i \sim \mathcal{N}(0, 1)$ are independently sampled, and $f(x_i)$ is simulated by the R package \texttt{RandomField}. Three hundred of the 6,000 observations were randomly selected as the training data $\dataset$, while the remaining 5,700 were reserved for test data. Figure \ref{fig:1d_illustration} illustrates how the patchwork kriging predictor changes for different $K$, relative to the global GP predictor and the regular local GP predictor with no continuity conditions across regions. As the number of regions ($K$) increases, the regular local GP predictor deviates more from the global GP predictor. The test prediction mean square error (MSE) for the regular local GP predictor at the 5,700 test locations is 0.0137 for $K=4$, 0.0269 for $K = 8$, 0.0594 for $K=16$, and 0.1268 for $K=32$. In comparison, patchwork kriging substantially improves the test MSE to 0.0072 for $K=4$,  0.0123 for $K = 8$, 0.0141 for $K=16$, and 0.0301 for $K=32$. 

\begin{figure}[h]
        \centering
                \includegraphics[width=\textwidth]{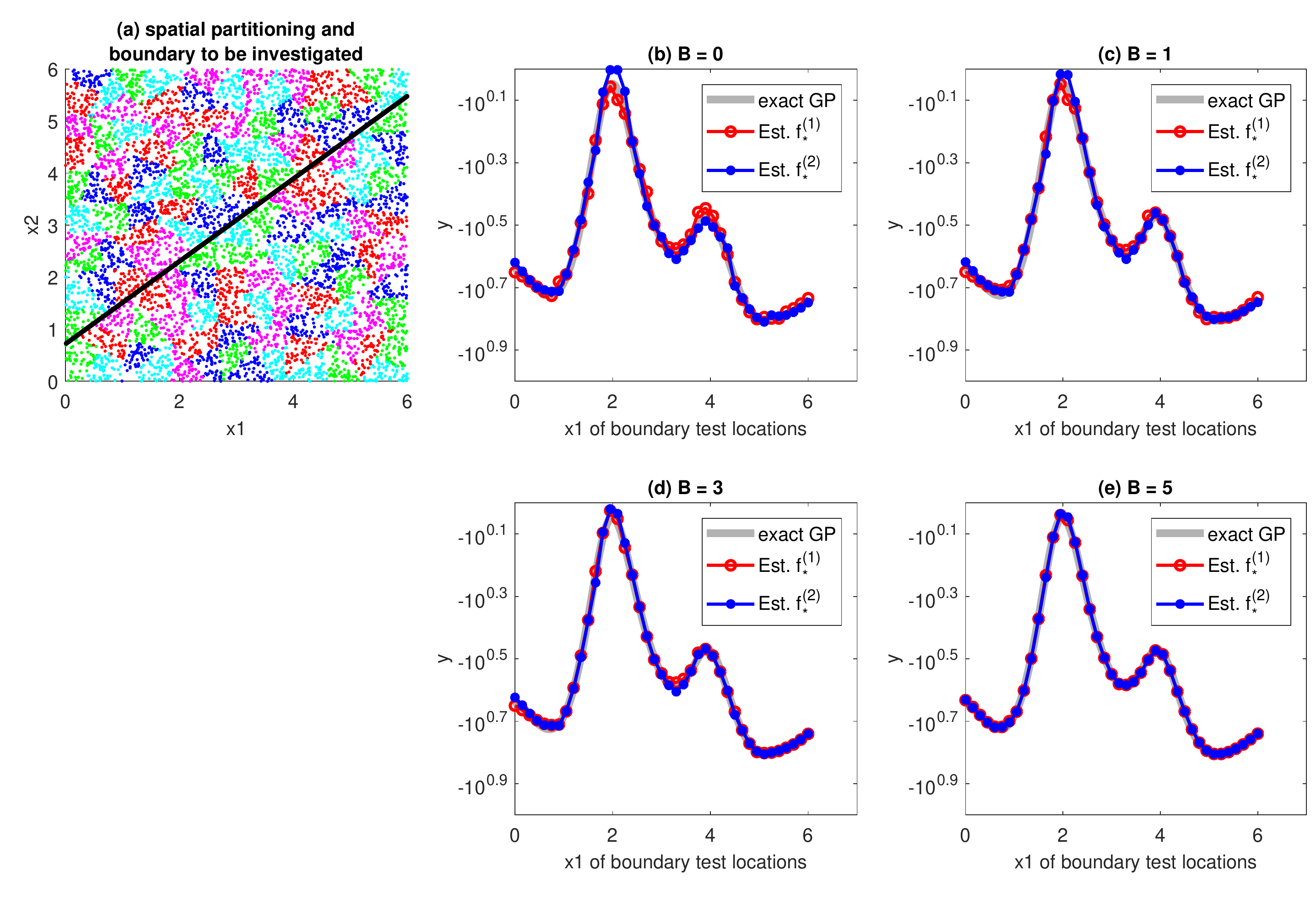}
        \caption{Comparison of the patchwork kriging mean predictions of two local models over interfacial boundaries. Panel (a) shows how the entire regression domain was spatially partitioned into 128 local regions, which are distinguished by their colors. The black solid line cutting through the entire space is the interfacial boundary that we selected to study the behavior of the patchwork kriging at interfacial boundaries. Panels (b), (c), (d) and (e) compare the patchwork kriging mean predictions of the neighboring local models when $B$ = 0, 1, 3, and 5. In the panels, the horizontal axes represent the x1 coordinates of test locations on the solid interfacial boundary line shown in panel (a). $f^{(1)}_*$ and $f^{(2)}_*$ denote the mean predictions of the two local models on each side of the solid boundary line. As $B$ increases, the two local predictions converge to each other, and the converged values are very close to the benchmark predictor achieved using the true GP model globally.}
                 \label{fig:sim2d_boundary_est}
 \end{figure}
We also generated a synthetic dataset in 2-d using the R package \texttt{RandomField}, and we denote this dataset by \texttt{synthetic-2d}. \texttt{synthetic-2d} consists of 8,000 noisy observations from a zero-mean Gaussian process with the exponential covariance function of $c(x_i, x_j) = 10\exp(-||x_i-x_j||_2)$,
\begin{equation*}
y_i = f(x_i) + \epsilon_i \quad \mbox{ for } i = 1,\ldots, 8000,
\end{equation*} 
where $x_i \sim \mbox{Uniform}([0, 6]\times[0, 6])$ and $\epsilon_i \sim \mathcal{N}(0, 1)$ were independently sampled. We used the dataset to illustrate how two local GP models for two neighboring local regions change as $B$ changes. We first partitioned the dataset into 128 local regions as shown in Figure \ref{fig:sim2d_boundary_est}-(a). For evaluation purposes, we considered test points that fell on the boundary cutting the entire regression domain into two (indicated by the black solid line in Figure \ref{fig:sim2d_boundary_est}-(a)), and sampled 201 test points uniformly over this boundary; the test locations do not coincide with the locations that pseudo observations placed. For each point, we get two mean predictions from the two local patchwork kriging models that straddle the boundary at that point. We compared the two mean predictions to each other for different choices of $B$ and also compared them with the optimal global GP prediction, i.e., the prediction using the true GP covariance function and the entire dataset globally without spatial partitioning.  Figure \ref{fig:sim2d_boundary_est} shows the comparison. When $B=0$, the two local models exhibited significant differences in their mean predictions. The differences decreased as $B$ increased, and became negligible when $B \ge 5$. The mean predictions were also very close to the exact GP predictions. The similar results were observed in different simulated examples, which will be discussed 

\section{Evaluation with Simulated Examples} \label{sec:simstudy}
In this section, we use simulation datasets to understand how the patchwork kriging behaves under different input dimensions, covariance ranges and choices of $K$ and $B$.  

\subsection{Datasets and Evaluation Criteria} \label{sec:simdata set}
\begin{figure}[t]
    \centering
            \includegraphics[width=\textwidth]{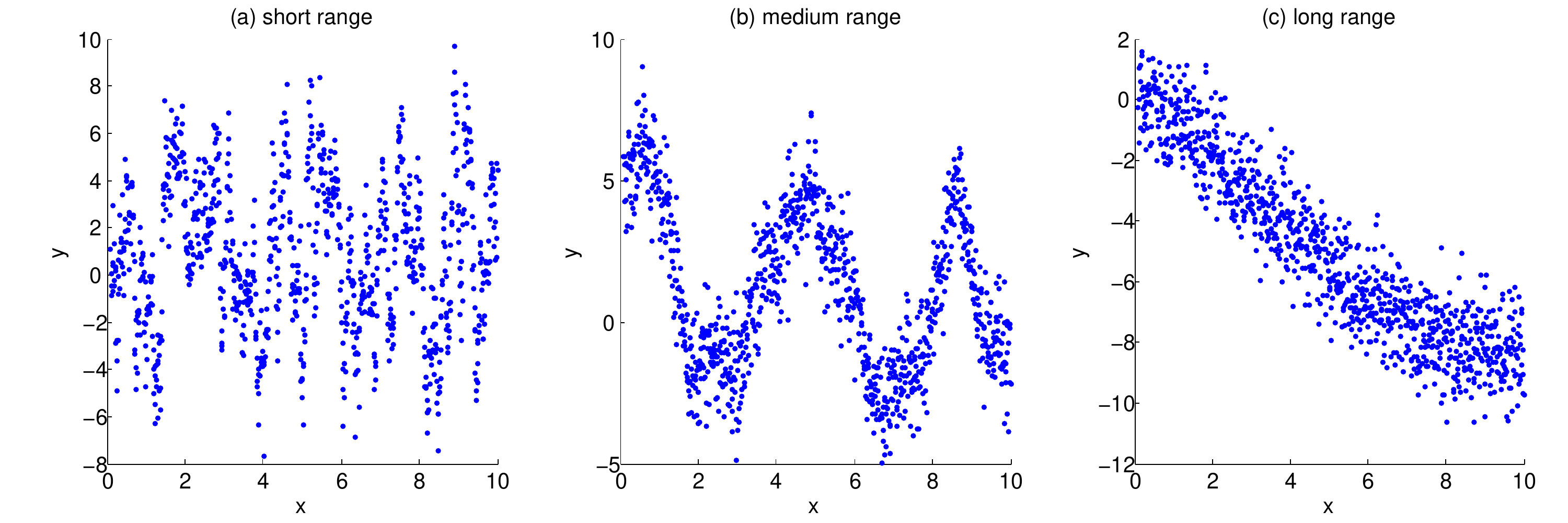}
    \caption{Illustrative Data with Short-range, Med-range and Long-range Covariances}
    \label{fig:data_range}
\end{figure}

Simulation datasets are sampled from a Gaussian process with the squared exponential covariance function, 
\begin{equation}
c(x_i, x_j) = \tau \exp\left(-\frac{(x_i-x_j)^T(x_i - x_j)}{2\rho^2}\right) \mbox{ for } x_i, x_j \in \mathbb{R}^d,
\end{equation}
where $\tau > 0$ is the scale parameter, and $\rho > 0$ determines the range of covariance. We randomly sampled 10,000 pairs of $x_i$ and $y_i$. Each $x_i$ is uniformly from $[0, 10]^d$, and then evaluate the sample covariance matrix with $\tau$ and $\rho$ for the 10,000 sampled inputs, $\M{C}_{\tau,\rho}$. All $y_i$'s are jointly sampled from $\mathcal{N}(\V{0}, \sigma^2 \M{I} + \M{C}_{\tau,\rho})$. We fixed $\tau=10$ and $\sigma^2 = 1$, but chose $\rho$ to 0.1 (short range), 1 (med range) or 10 (long range) to simulate datasets having different covariance ranges; see Figure \ref{fig:data_range} for illustrating simulated datasets for one dimensional input. In addition, the input dimension $d$ was varied over $\{2, 5, 10, 100\}$. In total, we considered 12 different combinations of different $\rho$ and $d$ values. For each combination, we drew 50 datasets, so there are 600 datasets in total.

For each of the datasets, we ran the patchwork kriging with different choices of $K \in \{16, 32, 64, 128, 256\}$ and $B \in \{0, 1, 2, 3, 5, 7, 10, 15, 20, 25\}$. For each run, we evaluate the computation time and prediction accuracy of patchwork kriging. For the prediction accuracy, we first computed the predictive mean of the optimal GP predictor (i.e., using the true exponential covariance function) at test locations, and we used these optimal prediction values as a benchmark against which to judge the accuracy of the patchwork kriging predictions. One thousand test locations are uniformly sampled from the interior of local regions, denoted by $\{x_t; t=1,...,T_I\}$, and 200 additional test locations were uniformly sampled from the boundaries between local regions, which are denoted by $\{x_t; t=T_I+1,...,T_I+T_B\}$. Let $\mu_t$ denote the estimated posterior predictive mean at location $x_t$, and let $\tilde{\mu}_t$ denote the benchmark predictive mean at the same location.  We measure three performance metrics for the mean predictions. The first two measures are the interior mean squared error (I-MSE) and the boundary mean squared error (B-MSE)
\begin{equation}
\textrm{I-MSE} = \frac{1}{T_I} \sum_{t=1}^{T_I} (\tilde{\mu}_t - \mu_t)^2, \textrm{ and B-MSE} = \frac{1}{T_B} \sum_{t=T_I+1}^{T_I+T_B} (\tilde{\mu}_t - \mu_t)^2,
\end{equation}
which measure the average accuracy of the mean prediction inside local regions and on the boundary of local regions. For each boundary point in $\{x_t; t=T_I+1,...,T_I+T_B\}$, we get two mean predictions from the two local patchwork kriging models that straddle the boundary at that point. In the B-MSE calculation, we took one of the two predictions following the rule: when $x_* \in \Gamma_{kl}$, choose the prediction for $f^{(k)}_*$ if $k < l$. Please note that when a test location is at a corner where three or more local regions meet, we do have more than two predictions, which did not happen in all of our testing scenarios. We also evaluated the squared difference of the two mean predictions for each of 200 boundary points, and the mean squared mismatch (MSM) was defined as the average of the squared differences. We also measured the three performance metrics for the variance predictions, which were named `I-MSE($\sigma^2$)', `B-MSE($\sigma^2$)' and `MSM($\sigma^2$)' respectively.

\subsection{Analysis of the Outcomes and Choices of $K$ and $B$} \label{sec:KBstudy}
Figures \ref{fig:MSE} show the I-MSE, B-MSE and MSM performance of the patchwork kriging for different covariance ranges and different choices of $K$ and $B$ when $d=100$, and Appendix C contains the plots of all six performance metrics for all simulation configurations. All of the performance metrics have shown the similar patterns:

\begin{figure}[t]
    \centering
            \includegraphics[width=\textwidth]{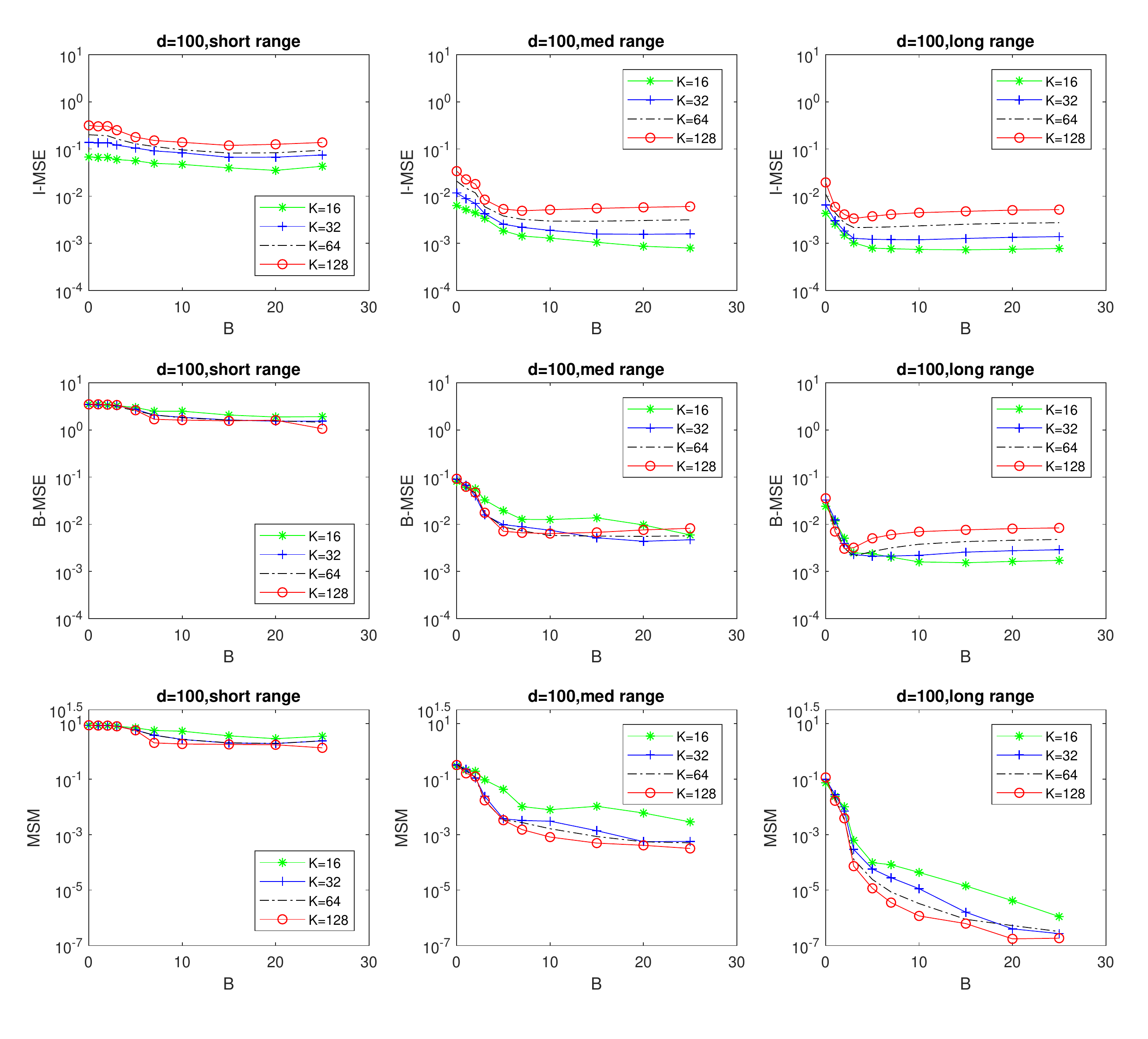}
    \caption{Performance of Patchwork Kriging for Simulated Cases with 100 Input Dimensions.}
    \label{fig:MSE}
\end{figure}

\begin{itemize}
\item \textbf{Covariance Ranges:} All of the performance metrics became negligibly small for medium-range and long-range covariances with large $B$. This implies that the patchwork kriging approximates the full GP very well for medium-range and longer-range covariances; please see Appendix for detailed plots. This result is opposite to our initial expectation that local-based approaches would have some deviations from the full GP for long-range covariances. As long as the underlying covariance is stationary, the proposed approach works well for long-range covariance cases. 
\item \textbf{Effect of $B$:} All of the metrics decrease in $B$ but does not change much for $B$ above 8 for medium-range and long-range covariances. However, when the covariance range is short, the improvement of the three metrics goes slower. This implies larger $B$ is required to achieve good accuracy for short-range covariances.
\item \textbf{Effect of $K$:} All of the metrics increase in $K$ when the other conditions are kept same. This is understandable, because the simulated data came from a stationary process. However, the effect of $K$ on the three metrics was relatively small when $B > 7$ and covariance ranges are medium or long. Since the computational complexity of the proposed method decreases with increase of $K$, choosing a large $K$ with $B > 7$ could be a computationally economic option with good prediction accuracy. See our computation time analysis below for an additional discussion on the choices of $K$ and $B$.
\item \textbf{Boundary Consistency:} Both of the MSM and MSM($\sigma^2$) goes to zero as $B$ increases for medium and long-range covariances. This implies that if data change smoothly, the patchwork idea does not only guarantee the same predictions on the locations pseudo data placed but also gives the same predictions over the entire inter-domain boundaries. 
\end{itemize}

\begin{figure}[t]
    \centering
            \includegraphics[width=0.7\textwidth]{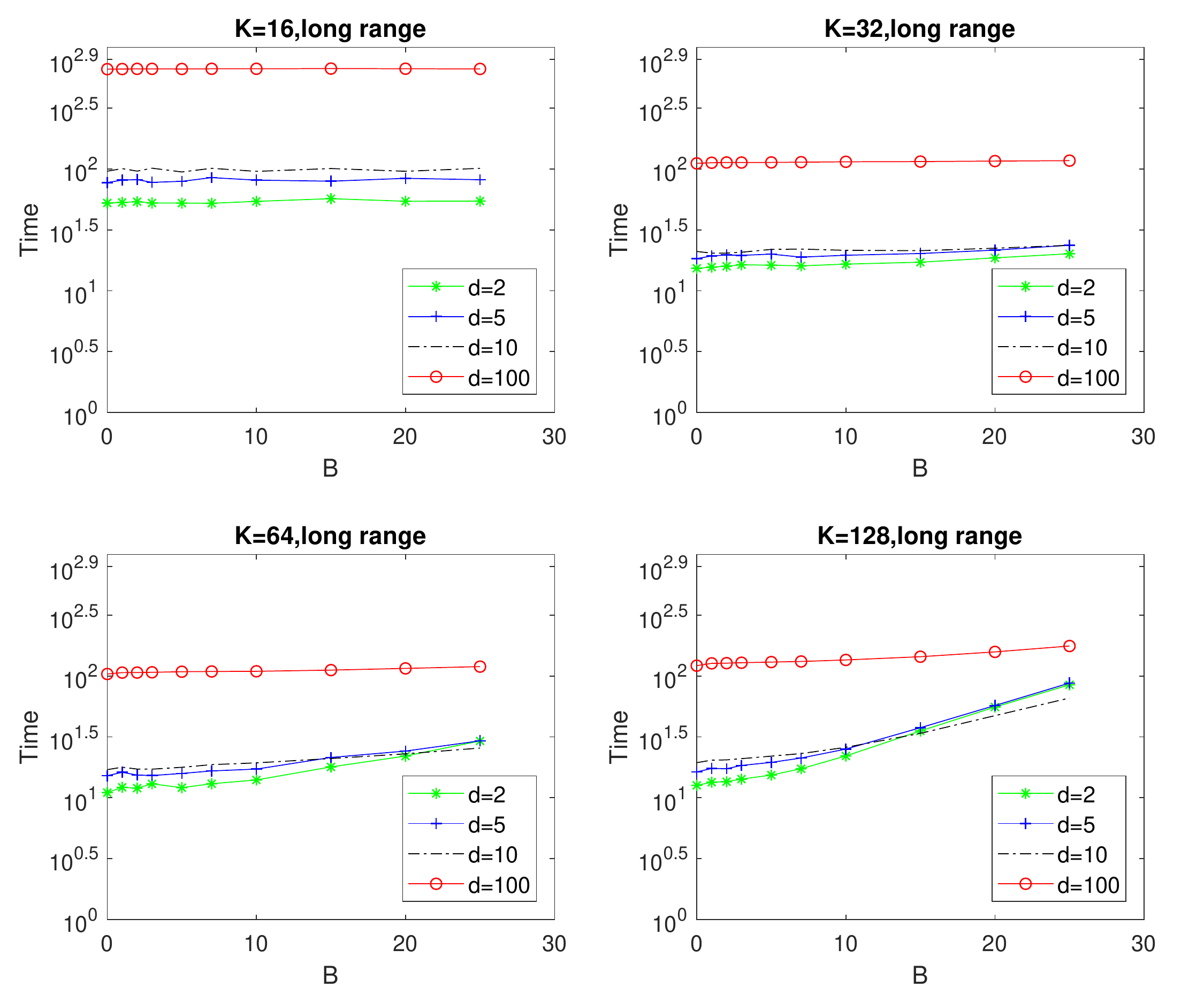}
    \caption{Summary of Total Computation Times for Simulated Cases.}
    \label{fig:TIME}
\end{figure}

Figure \ref{fig:TIME} summarizes the total computation time of the patchwork kriging for different configuration.
\begin{itemize}
\item It appears that the input dimension is not a determining factor of the time if the input dimension $d \le 10$, but for $d > 10$, it became a major factor to affect the time. As we discussed in Section \ref{sec:complexity}, the computational complexity of the patchwork kriging is $O(N^3/K^2 + d_f^3B^3K)$. When data are uniformly located over the regression domain, $d_f \propto d$ and the computation of the patchwork kriging is scaling proportionally to $d^3$. 
\item When $d \le 10$, the deciding factor for the computation time was $K$. In general, larger $K$ gave shorter computation times.
\item With larger $K$, $B$ becomes more influential to the total computation time. This is due to the increase of the second term in the overall computational complexity, $O(N^3/K^2 + d_f^3B^3K)$. 
\item To keep the total computation time lower, both of $N/K$ and $d_fB$ should be kept lower. On the other hand, $N/K$ and $d_fB$ cannot be too small due to degradation of prediction accuracy with a large $K$ and a small $B$. 
\end{itemize}
Our numerical studies suggest to choose $K$ so that $N/K$ be in between 200 and 600 and then choose $B$ so that $d_fB$ is in between $15$ and $400$ to balance off the computation and prediction accuracy; these were based on all of the simulation cases presented in this section as well as the six real data studies that will be presented in the next section. In order to keep $d_fB \le 400$ for efficient computation and $B \ge 7$ for prediction accuracy, $d_f\le 400/7$. Therefore, the proposed approach would benefit more for $d_f \le 55$. However, the proposed approach still worked better than some existing approaches for the simulated cases with $d=100$; see the numerical results in Sections \ref{sec:comp_fitc} and \ref{sec:sim_rbcm}.

\subsection{Comparison to a Global Approximation Method} \label{sec:comp_fitc}
\begin{figure}[h]
    \centering
            \includegraphics[width=\textwidth]{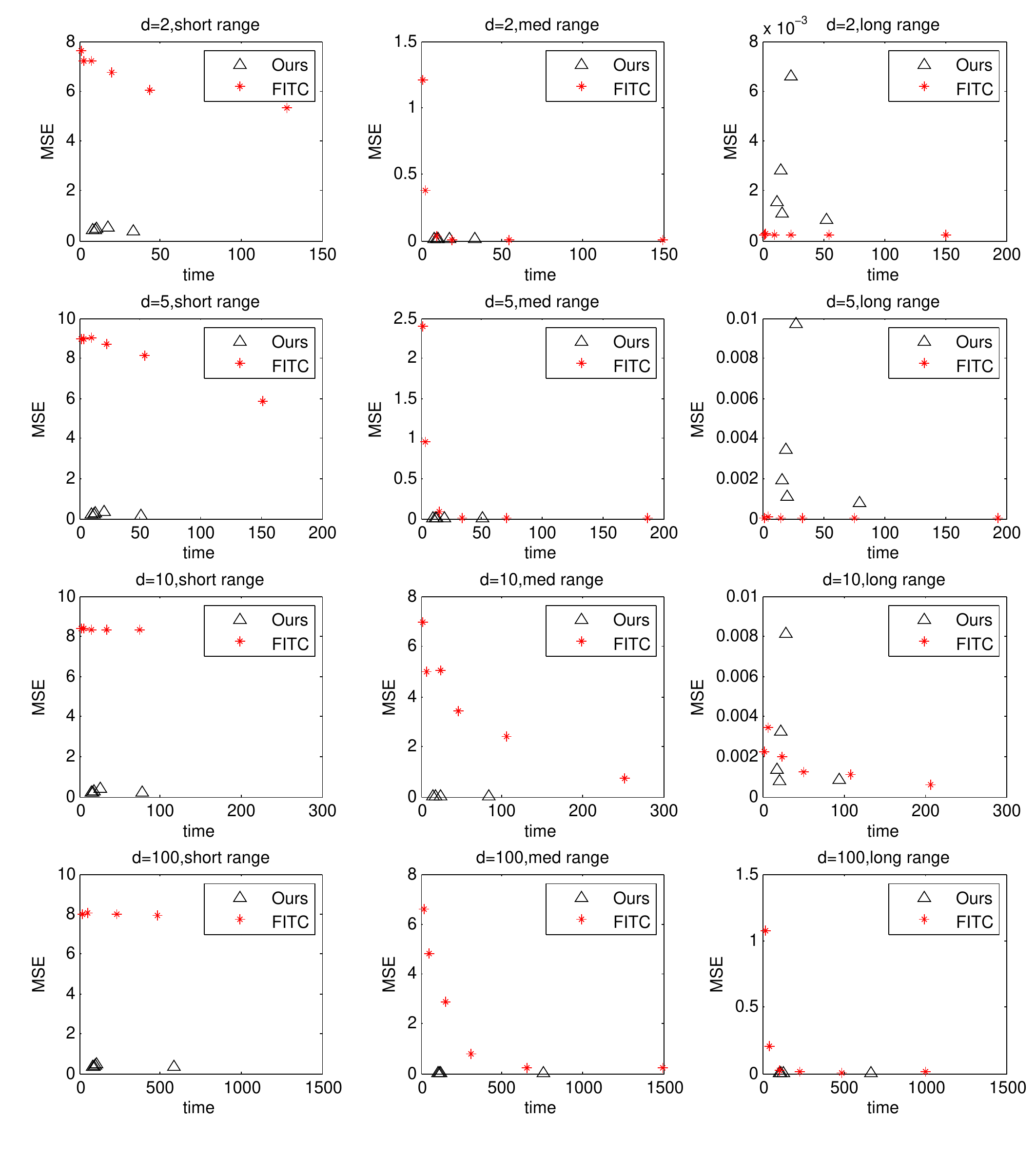}
    \caption{Comparison of Total Computation Times vs. MSE for Simulated Cases; triangles and stars represent the results of the patchwork kriging and FITC respectively. }
        \label{fig:COMP}
\end{figure}

We also used the simulated cases to compare the patchwork kriging to a global GP approximation method, the Fully Independent
Training Conditional (FITC) algorithm \citep[FITC]{snelson:06}. We decided to compare ours with the global GP approximation method because we thought that the global GP approximation would work better when stationary covariances are used. For the patchwork kriging, we fixed $B=7$ and varied $K \in \{32, 64, 128, 256\}$. For the FITC, the total number of pseudo inputs was varied over $\{16, 32, 64, 128, 256, 512\}$. The computation times versus MSE of the mean prediction were compared for different input dimensions and covariance ranges. Figure \ref{fig:COMP} summarizes the outcome. The patchwork kriging outperformed the FITC with significant performance gaps for all short covariance and medium range covariance cases. For long range covariance cases, the FITC performed better when $d < 10$, but the patchwork kriging performed comparably when $d \ge 10$. 

The performance gap in between the FITC and the patchwork kriging can be explained by a more efficient computation of the patchwork kriging. Both of the FITC and patchwork kriging use pseudo inputs. Their accuracies depend on the total number of pseudo inputs used. When $Q$ pseudo inputs were applied for both of FITC and the patchwork kriging, the computation of the FITC involves the inversion of a $Q \times Q$ dense matrix, while the computation of the patchwork kriging involves the inversion of the sparse matrix of the same size that corresponds to equation \eqref{eq:Cinv}. Therefore, when comparable computation times were invested, the patchwork kriging could place more pseudo inputs than the FITC, so it can give better accuracy. In addition, the locations of the pseudo inputs in the FITC need to be learned together with covariance hyperparameters, and the increase in the number of pseudo inputs would increase the computation time for hyperparameter learning. 

\subsection{Comparison to Local Approximation Methods} \label{sec:sim_rbcm}
\begin{figure}[t]
    \centering
            \includegraphics[width=0.95\textwidth]{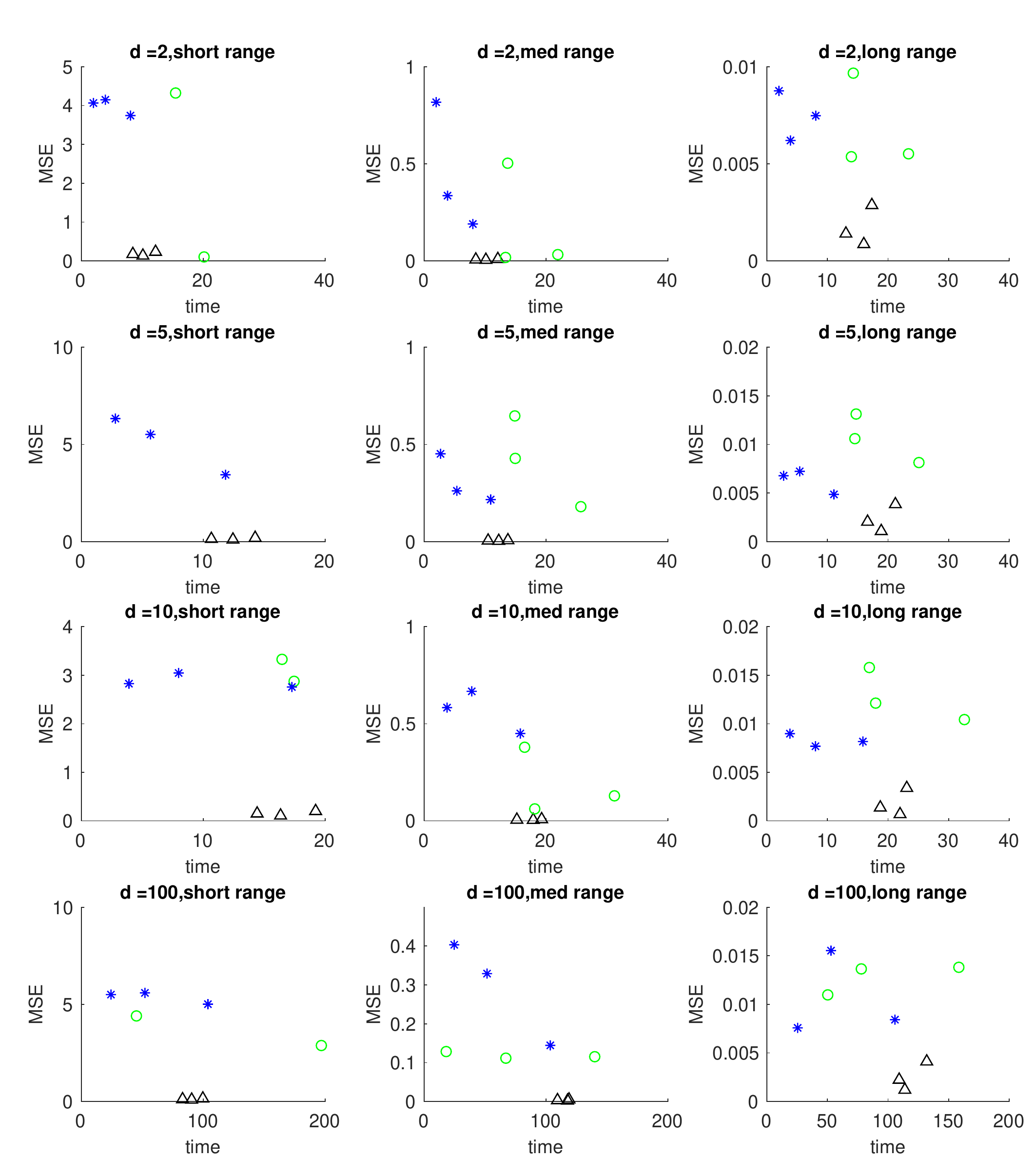}
    \caption{Comparison of Total Computation Times vs. MSE for Simulated Cases; triangles, stars and circles represent data for the patchwork kriging, PIC and RBCM respectively. The number of circles are not always same because PIC could not produce outcomes for some simulated cases due to singularity in numerical inversion.}
        \label{fig:COMP2}
\end{figure}
We also used the simulated cases to compare the patchwork kriging to two local GP approximation methods, a robust Bayesian committee machine \citep[RBCM]{deisenroth2015}, and a partially independent conditional approach \citep[PIC]{snelson:07}. For the patchwork kriging, we fixed $B=7$ and varied $K \in \{32, 64, 128\}$. For RBCM, we used $K \in \{32, 64,128\}$. For PIC, we used $K \in \{32,64,128\}$ with the total number of pseudo inputs fixed to $128$. Figure \ref{fig:COMP2} summarizes the comparison of MSE performance. The patchwork kriging performed very competitively for all simulated cases. The significant increase of computation time for the input dimension more than 10 was observed for all of the compared methods. 

\section{Evaluation with Real Data} \label{sec:validation}
In this section, we use five real datasets to evaluate the patchwork kriging and compare it with the state-of-the-art, including \citep[PGP]{park:16}, a Gaussian Markov random field approximation \citep[GMRF]{lindgren2011}, a robust Bayesian committee machine \citep[RBCM]{deisenroth2015}, and a partially independent conditional approach \citep[PIC]{snelson:07}. The comparison with one additional dataset is presented in Appendix D. 

\subsection{Datasets and Evaluation Criteria} \label{sec:data set}
We considered five real datasets: two spatial datasets in 2-d with different spatial distributions of observations, one additional spatial dataset with a very large data size, and three higher  dimensional datasets, one with 9-dimension, another with 21-dimension and the other with 8-dimension. 

The first spatial dataset, \texttt{TCO.L2}, 182,591 observations collected by the
NIMBUS-7/TOMS satellite, which measures the total column of ozone over the globe on Oct 1 1988. Two predictors represent the longitude and latitude of a measurement location, while the corresponding independent variable represents the measurement at the location. The observations are uniformly spread over the range of the two predictors. The main analysis objective with this dataset is to predict the total column of ozone at unobserved locations. 

The second dataset, \texttt{ICETHICK}, contains ice thickness measurements at 32,481 locations on the western Antarctic ice sheet and is available at \url{http://nsidc.org/}. It has two predictors that represent the longitude and latitude of a measurement location, and the corresponding independent variable is the ice thickness measurement. The dataset has many sparse regions where there are very few observations. Regression analysis with this dataset would give the prediction of ice thickness at unobserved locations.

The third dataset, \texttt{PROTEIN}, has nine input variables that describe the tertiary structure of proteins and one independent variable that describes the physiochemical property of proteins. These data, which are available at \url{https://archive.ics.uci.edu/ml/datasets}, consist of 45,730 observations. Like typical high dimensional datasets, the measurements are embedded on a low dimensional subspace of the entire domain.  This dataset can be studied to relate the structure of a protein with the physiochemical property of the protein for predicting the property from the structure.

The fourth dataset, \texttt{SARCOS}, contains measurements from a seven degrees-of-freedom SARCOS anthropomorphic robot arm. There are 21 predictors that describe the positions, moving velocities and accelerations of seven joints of the robot arm, and the seven response variables are the corresponding torques at the seven joints. We only use the first response variable for this numerical study. The dataset, which is available at \url{http://www.gaussianprocess.org/gpml/data/}, contains 44,484 training observations and 4,449 test observations. The main objective of the regression analysis is to predict one of the joint torques in a robot arm when the values of the predictors are available.

The last dataset, \texttt{FLIGHT}, consists of 800,000 flight records randomly selected from the database available at \texttt{http://stat-computing.org/dataexpo/2009/}. The same size subset of the database was used as a benchmark dataset in literature \citep{Hensman13}. Following the use in the literature, we used 8 predictors that include the age of the aircraft, distance that needs to be covered, airtime, departure time, arrival time, day of the week, day of the month and month, and the response variable is the arrival time delay. This dataset was studied to predict the flight delay time when the predictors are given.

Using the five datasets, we compare the computation time and prediction accuracy 
of patchwork kriging with other methods. We randomly split each dataset into a training
set containing 90\% of the total observations and a test set containing
the remaining 10\% of the observations. To compare the computational
efficiency of methods, we measure total computation times. For comparison of prediction accuracy, we measure two performance metrics on the test data, denoted by $\{(x_t, y_t): t=1,\dots,T\}$, where $T$ is the size
of the test set. Let $\mu_t$ and $\sigma^2_t$ denote the estimated posterior predictive mean and variance at location $x_t$; when the testing location $x_t$ is in the domain boundary $ \Gamma_{kl}$, we may have two predictions, one for $f^{(k)}(x_t)$ and the other for $f^{(l)}(x_t)$, for which we choose one for $f^{(k)}(x_t)$ if $k < l$. Please note that when a test location is at a corner where three or more local regions meet, we do have more than two predictions, which did not happen in all of our testing scenarios. We also evaluated the squared  The first measure is the mean squared error (MSE)
\begin{equation}
\textrm{MSE} = \frac{1}{T} \sum_{t=1}^T (y_t - \mu_t)^2,
\end{equation}
which measures the accuracy of the mean prediction $\mu_t$ at location $x_t$.
The second measure is the negative log predictive density (NLPD)
\begin{equation}
\textrm{NLPD} = \frac{1}{T} \sum_{t=1}^T\left[ \frac{(y_t - \mu_t)^2}{2\sigma_t^2} + \frac{1}{2} \log (2\pi \sigma_t^2) \right].
\end{equation}
The NLPD quantifies the degree of fitness of the estimated predictive distribution $\mathcal{N}(\mu_t, \sigma_t^2)$ for the test data. These two criteria are used broadly in the GP regression literature. A smaller value of MSE or NLPD indicates better performance.  All numerical experiments were performed on a desktop computer with Intel Xeon Processor W3520 and 6GB RAM.

The comparison was made in between our method and the state-of-the-art previously listed. Note that the PGP and the GMRF approaches cannot be applied for more than two input dimensions, and so were only compared for the three spatial datasets. We tried two covariance functions, a squared exponential covariance function and an exponential covariance function. Note that the PIC method does not work with an exponential covariance function because learning the pseudo inputs for the PIC method requires the derivative of a covariance function but an exponential covariance function is not differentiable. On the other hand, when an squared exponential covariance function is applied to the GMRF, the precision matrix construction is not straightforward. Therefore, we used a squared exponential covariance function for comparing the proposed approach with the PIC, RBCM, and PGP, while using an exponential covariance function for comparing it with the GMRF. For both of the cases, we assumed the same hyperparameters for local regions, and we used the entire training dataset to estimate the hyperparameters. 

We chose and applied different partitioning methods for the compared methods. The choice of the partitioning schemes for the patchwork kriging and PGP is restrictive because every local region needs to be simply connected to minimize the area of the boundaries between local regions, so we used the spatial tree. The GMRF comes with a mesh generation scheme instead of a partitioning scheme, and following the suggestion by the GMRF's authors, we used the voronoi-tessellation of training points for the mesh generation. We tested the k-means clustering and the spatial tree for PIC and RBCM, but the choice did not make much difference in their performance. The results reported in this paper were the ones with the k-means clustering. 

We tried different numbers of the local regions that partition an input domain, and the numbers of the local regions were ranged so that the numbers of observations per local region would be approximately in between 80 and 600 for the proposed approach. The numbers were similarly ranged for the other compared methods with some variations to have the computation times of all the compared methods comparable; note that we like to compare the prediction accuracies of the methods when the computation times spent are comparable. For patchwork kriging, the locations of pseudo observation were selected using the rule described in Section \ref{sec:partition}. For PIC, the locations were regarded as hyperparameters and were optimized using marginal likelihood maximization.

\subsection{Example 1: \texttt{TCO.L2} Dataset}
This dataset has two input dimensions, and the inputs of the data are densely distributed over a rectangular domain. For patchwork kriging, we varied $B\in\{3,5\}$ and  $K \in \{256,512,1024\}$. The prediction accuracy of the PGP did not depend on the number of local regions $K$, so we fixed $K=623$, while the number of finite element meshes per local region was varied from 5 to 25 with step size 5. For RBCM, we varied the number of local experts $K \in \{ 100, 150, 200, 250, 300, 600\}$. For PIC, $K$ was varied over $\{100, 200, 300, 400, 600\}$, and the total number of pseudo inputs was also varied over $\{50, 70, 80, 100, 150, 200, 300\}$.

\begin{figure}[t]
    \centering
            \includegraphics[width=0.9\textwidth]{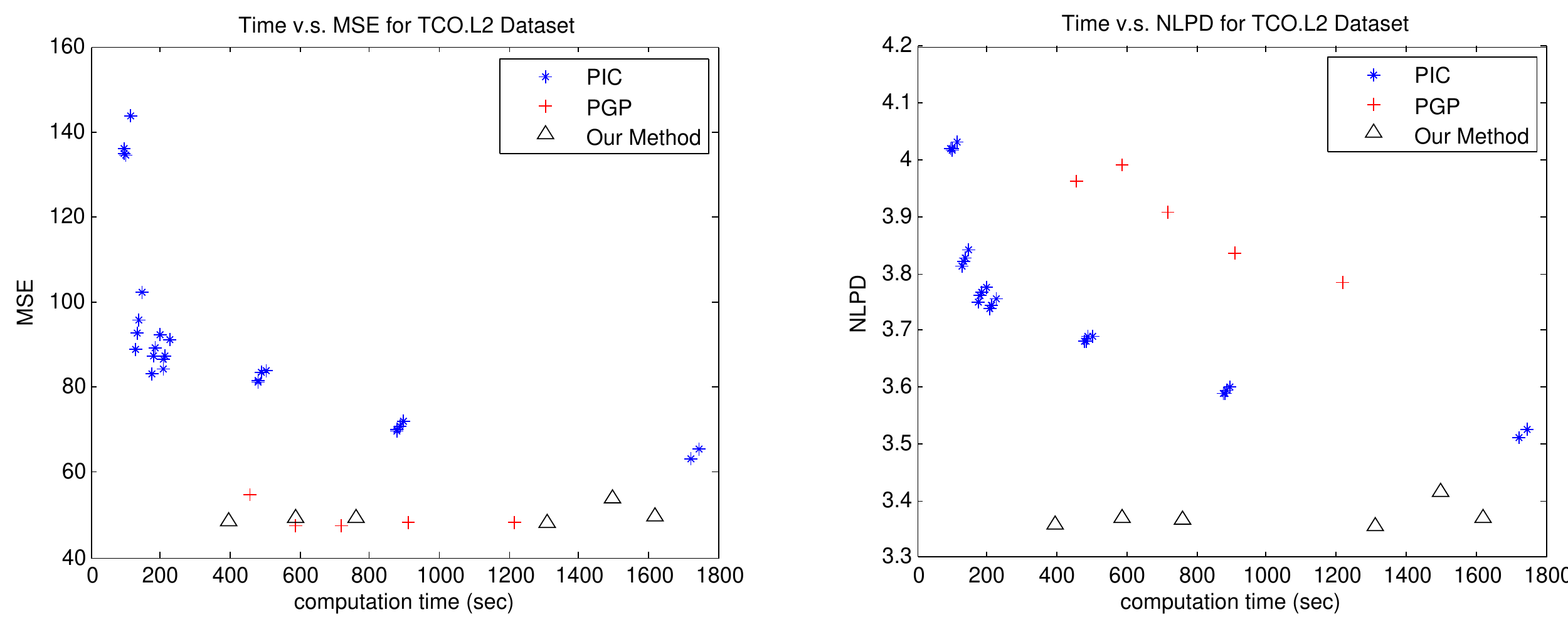}
             \includegraphics[width=0.9\textwidth]{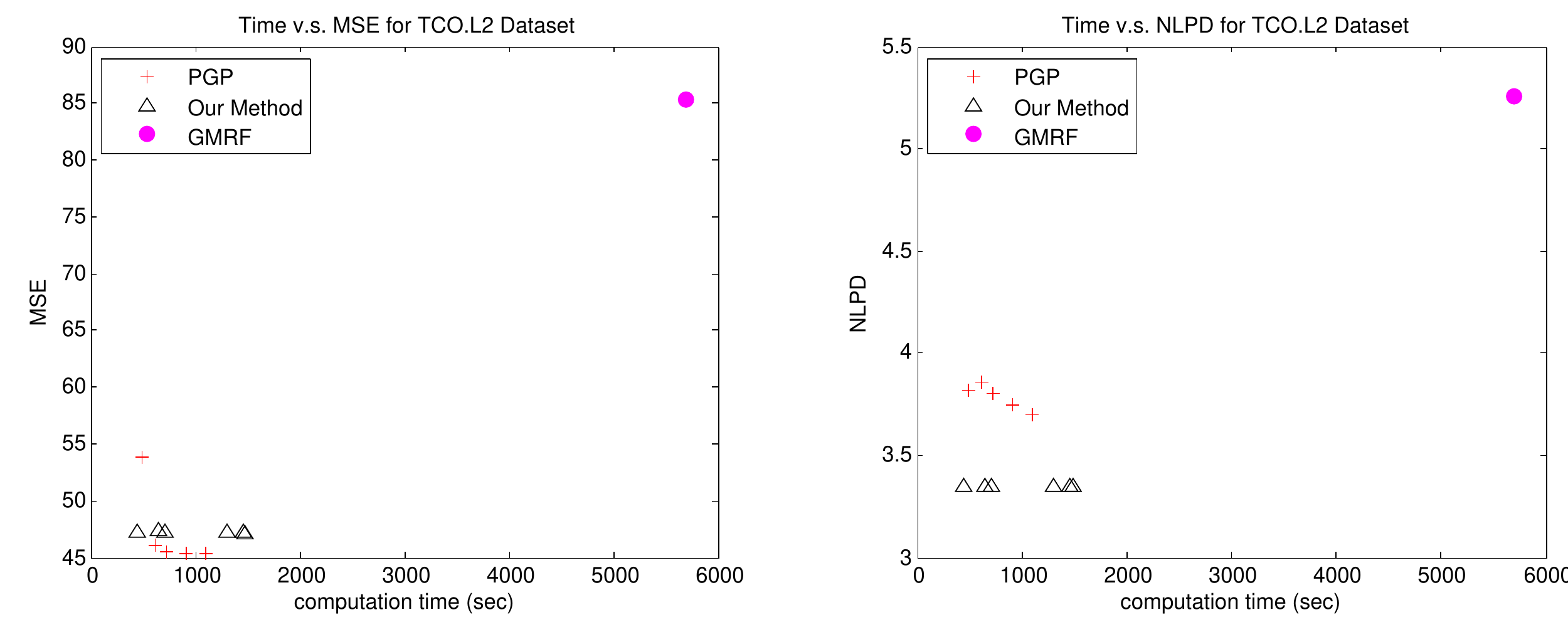}
    \caption{Prediction accuracy versus total computation time for the \texttt{TCO.L2} data.}
    \label{fig:result_tcoL2}
\end{figure}

\begin{figure}[t]
    \centering
            \includegraphics[width=0.9\textwidth]{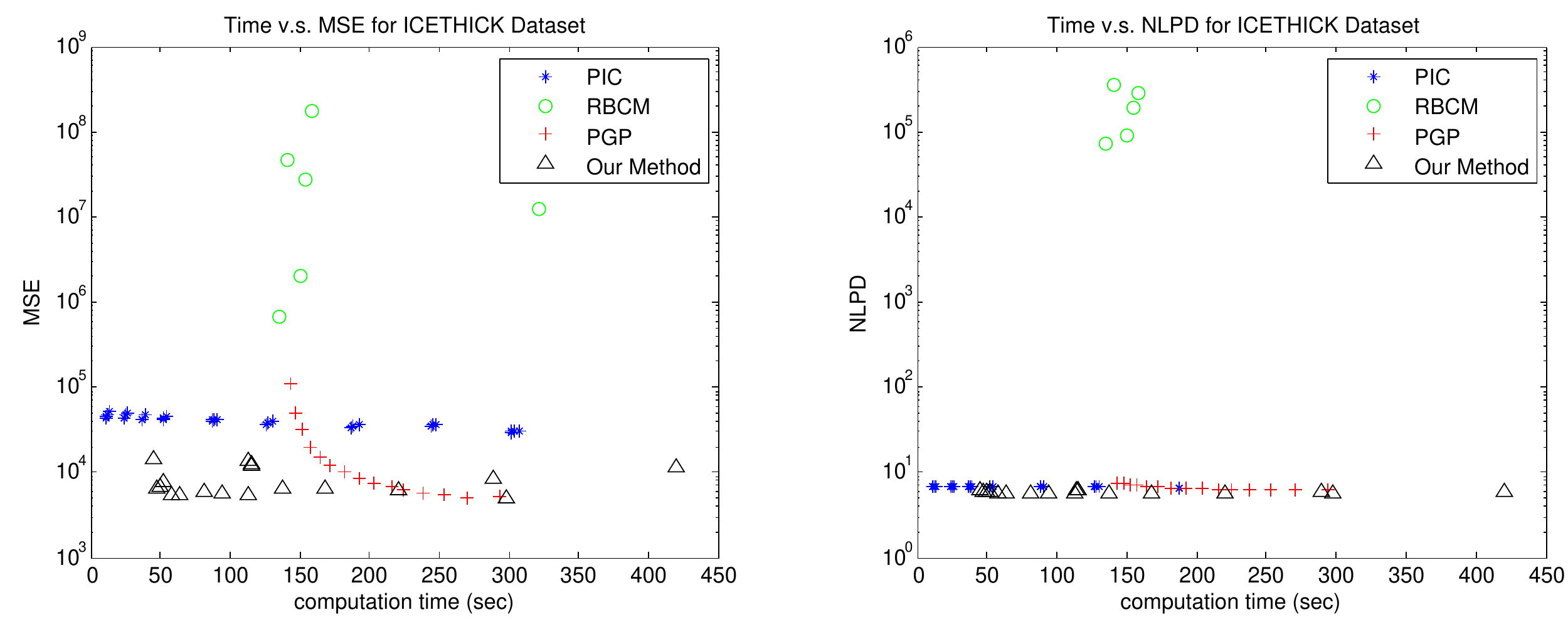}
            \includegraphics[width=0.9\textwidth]{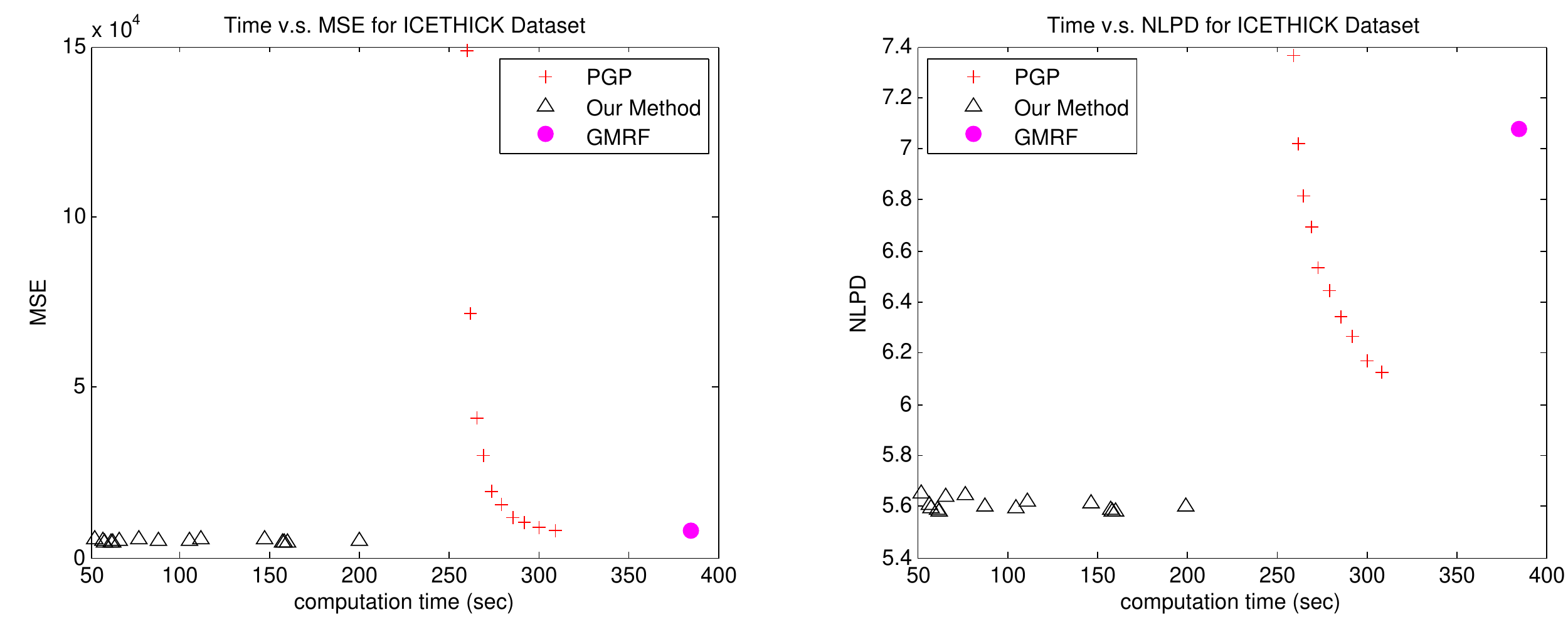}
    \caption{Prediction accuracy versus total computation time for the \texttt{ICETHICK} data.  A squared exponential covariance function was used for the results in the top panel, while an exponential covariance function was used for the results in the bottom panel. }
    \label{fig:result_icethick}
\end{figure}
Figure \ref{fig:result_tcoL2} shows the main results. The shortest computation time of RBCM (2319 seconds) was much longer than the longest time of the other compared methods, while its MSE was not competitive as well. Therefore we did not plot its results in the figure. For both of the square exponential and the exponential covariance functions, our approach and the PGP approach had comparable MSE. However, our approach significantly outperformed the PGP and PIC approaches in terms of the NLPD. This implies that our approach provides more accurate variance estimations.

\subsection{Example 2: \texttt{ICETHICK} Dataset}
One characteristic of this dataset is the presence of many spatial voids where there are no or very little data points. For patchwork kriging, we varied $B \in \{3,5,7\}$ and $K \in \{64,128,256,512,1024\}$.  For the PGP, we used $K = 47$, while the number of finite element meshes per local region was varied from 5 to 40 with step size 5. For RBCM, we varied the number of local experts $K \in \{ 50, 100, 150, 200, 250, 300\}$. For PIC, $M$ was varied over $\{50, 100, 150, 200\}$, and the total number of pseudo inputs was also varied over $\{50, 100, 150, 200, 300, 400, 500, 600, 700\}$.

Figure \ref{fig:result_icethick} compares the MSE and NLPD performance of the methods. Again, the PGP approach and the proposed approach outperformed the other methods, and the proposed approach achieved the best accuracy with much less computation time than the PGP approach. In addition, the proposed approach uniformly outperformed the other methods in terms of the NLPD. In other words, the proposed approach gives a predictive distribution that better fits the test data.

\begin{figure}[t]
    \centering
            \includegraphics[width=\textwidth]{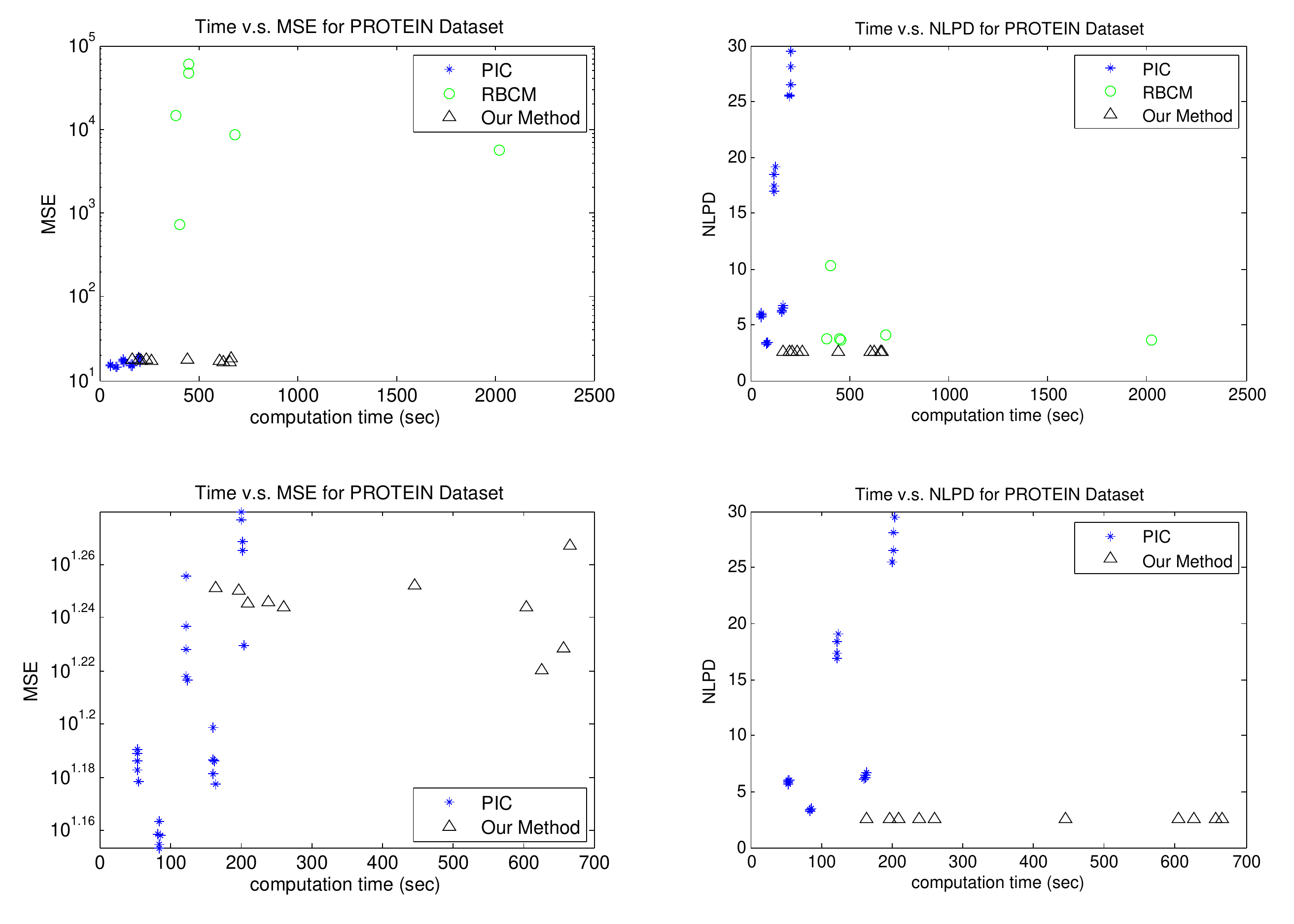}
    \caption{Prediction accuracy versus total computation time for the \texttt{PROTEIN} data.  The upper panel compares all three methods. Since the performance of the PIC and our method was very close, the bottom panel provides a closer look of the PIC and our method.}
    \label{fig:result_protein}
\end{figure} 

\subsection{Example 3: \texttt{PROTEIN} Dataset} 
Different from the previous datasets, this dataset features nine input variables. We will use this dataset to see how the proposed approach works for input dimension more than two. For the patchwork kriging, we varied $B \in \{2, 3, 4\}$ and varied $K \in \{64, 128, 256\}$; we have not included the results for larger $B$ because a larger $B$ increased the computation times of the patchwork kriging to a range incomparable to those of the other algorithms.  The PGP and GMRF approaches do not work with input dimensions more than two, and so were not included in this comparison. For RBCM, we varied the number of local experts $K \in \{100, 150, 200, 250, 300\}$. For PIC, $K$ was varied over $\{100, 150, 200, 250, 300\}$, and the total number of pseudo inputs ($M$) was also varied over $\{100, 150, 200, 250, 300\}$. In this comparison, we used a squared exponential covariance function for all three methods.

\begin{figure}[t]
    \centering
            \includegraphics[width=\textwidth]{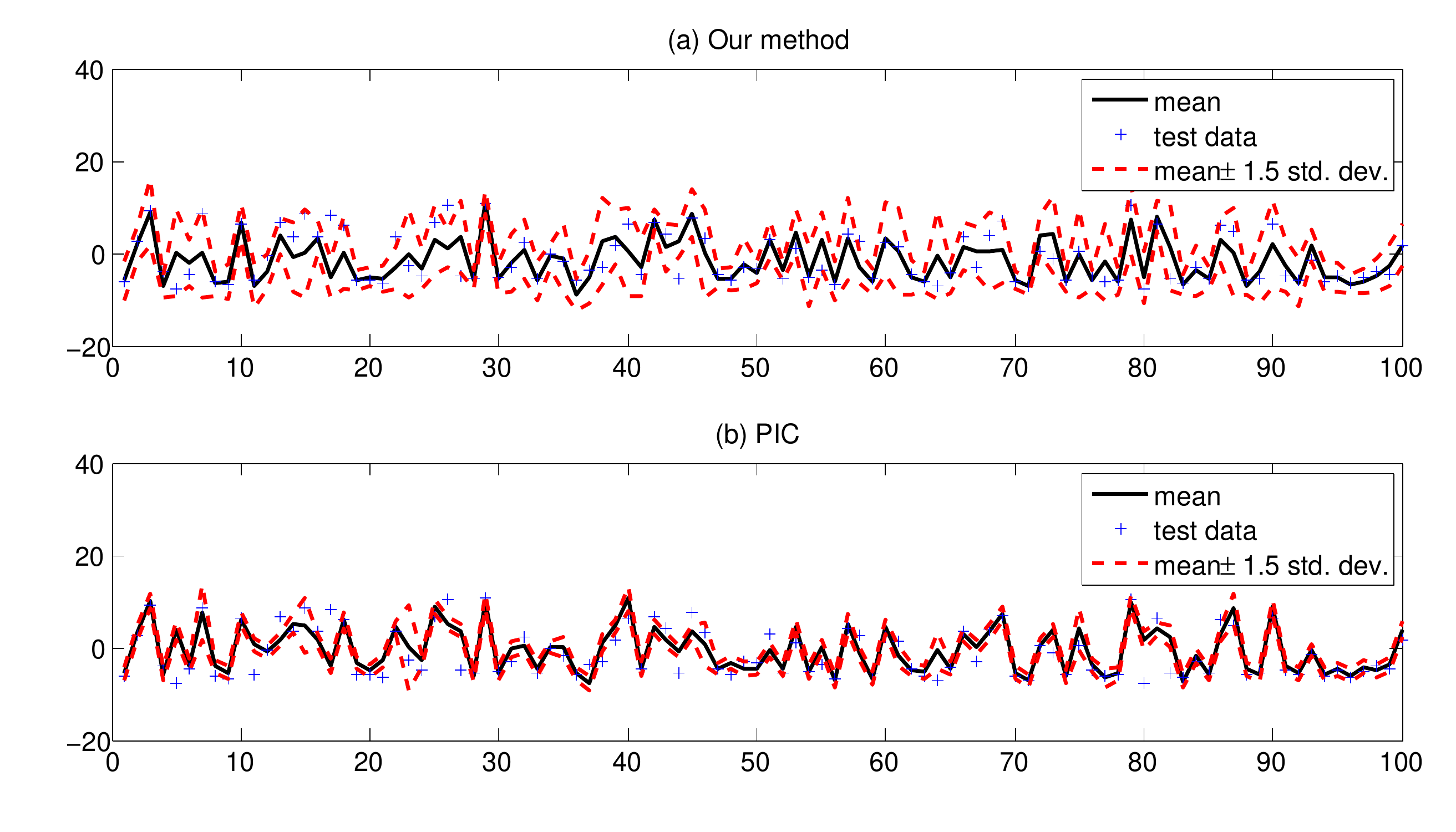}
    \caption{Comparison of the predictive distributions estimated by our method and by the PIC method for the \texttt{PROTEIN} data. }
    \label{fig:result_protein_int}
\end{figure}

Figure \ref{fig:result_protein} shows the main results. For this dataset, the PIC approach outperformed our method in terms of the MSE performance, providing more accurate mean predictions. On the other hand, our method provided better NLPD performance, which implies that the predictive distribution estimated by our method was better fit to test data than that of the PIC. Figure \ref{fig:result_protein_int} compares the predictive distributions estimated by the two methods. In the figure, the predicted mean $\pm 1.5$ predicted standard deviations was plotted for 100 randomly chosen test observations. The interval for the PIC was overly narrow and excluded many of the test response observations, while the interval for our method more appropriately covered the majority of data, which is reflected in the better NLPD performance for our method. The percentages of the 4,573 test observations falling within the intervals was 50.47\% for the PIC and 86.53\% for our method. Note that the probability of a standard normal random variable within $\pm 1.5\sigma$ is 86.64\%. Clearly, our method provides a better fit to the test data. 

\subsection{Example 4: \texttt{SARCOS} Dataset} 
This dataset has 21 input variables. For patchwork kriging, we varied $B \in \{3, 5, 7\}$, and we varied $K \in \{128, 256\}$. Again, the PGP and GMRF approaches do not work with high dimensional inputs, and so were not included in this comparison. For the RBCM approach, we varied the number of local experts $K \in \{100, 150, 200, 250, 300\}$. For PIC, $K$ was varied over $\{100, 150, 200, 250, 300\}$, and the total number of pseudo inputs ($M$) was also varied over $\{100, 150, 200, 250, 300\}$. In this comparison, we used a squared exponential covariance function for all three methods.

\begin{figure}[t]
    \centering
            \includegraphics[width=\textwidth]{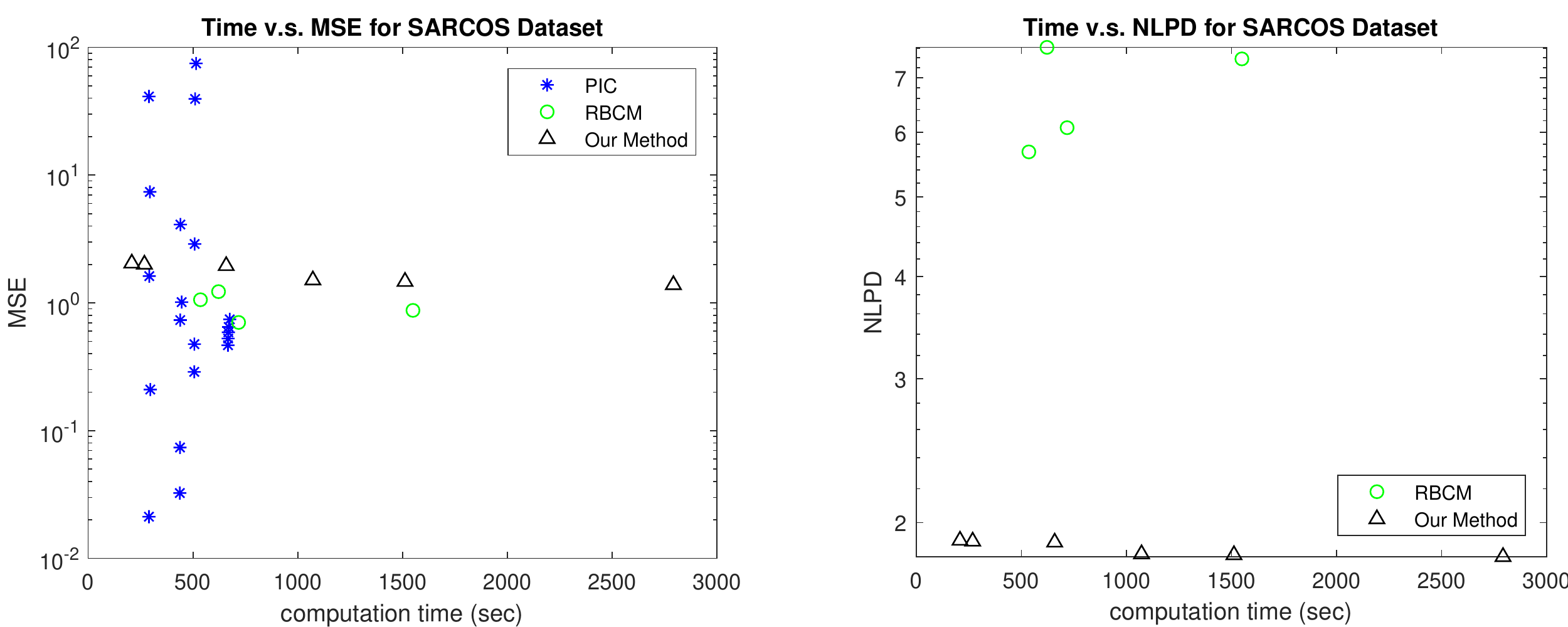}
    \caption{Prediction accuracy versus total computation time for the \texttt{SARCOS} data. A squared exponential covariance function was used. The PIC approach produced negative predictive variances, so its NLPD could not be computed. In the MSE plot, four triangles are supposed to show up. However, two of the four triangles are very closely located, so it looks like that there are only 3 triangles.}
    \label{fig:result_sarcos}
\end{figure} 

Figure \ref{fig:result_sarcos} summarizes the comparison of the MSEs and the NLPDs for the three methods. The MSE performances were comparable for all of the methods, while our approach provided a better fit to test data, which was evidenced by the smaller NLPD values of our approach. The PIC produced negative predictive variances for this dataset, so its NLPD values could not be calculated. In theory, the PIC approach should provide non-negative predictive variances with infinite precision. It evidently produced negative variances because of numerical errors. To be more specific, the numerical errors are mostly related to the inversion of covariance matrix of pseudo inputs. The condition number of the covariance matrix was very large, which incurred some round-off errors. Our patchwork kriging approach did not experience any such numerical issues in any of the examples, and it appears to be more numerically stable than the PIC approach.

\subsection{Example 5: \texttt{Flight Delays} Dataset}
This dataset has 800,000 records and eight input variables. For this dataset, due to memory limitation of our testing environment, we could not try various cases with different choices of tuning parameters. For patchwork kriging, we fixed $K = 1024$ and $B = 5$. For the RBCM, we set the number of local experts $K = 1024$. For PIC, we set $K = 1024$, and the total number of pseudo inputs $(M)$ was chosen to 1000 or 1500, and the further increase of $M$ gave an out-of-memory error in our testing environment. Note that PIC requires to precompute $K$ dense covariance matrices of size $N/K \times M$, which could be very large for this scale of $N$ and $M$. 

Table \ref{tbl:flight} summarizes the MSE and NLPD performance of our method, RBCM and PIC. The proposed approach gave a better MSE than the RBCM with less use of computation time. PIC showed very competitive computation performance, but its MSE was not as good as the MSE of our method. The increase of $M$ could improve the PIC's MSE performance, but the improvement was not big in between $M=1000$ and $M=1500$. 
\begin{table}
\centering
\begin{tabular}{c c c l}
\hline
Methods & MSE & NLPD & Computation Time (seconds) \\
\hline
Ours & 1188.4 & 4.8917  & 11729 \\
RBCM & 9790.2 & 10.5630 & 13218 \\
PIC ($M=1000$)  & 1494.8 & 5.0742  & 1624 \\
PIC ($M=1500$)  & 1492.7 & 5.0731  & 2094 \\
\hline
\end{tabular}
\caption{Comparison of MSE and NLPD performance for \texttt{Flight Delays} Dataset.}
\label{tbl:flight}
\end{table}

\subsection{Discussion}
Based on the numerical comparison results for the two spatial datasets and the three higher dimensional datasets, we can summarize the benefits of the proposed approach as threefold. First, it provides competitive or better mean prediction accuracy for both the spatial datasets and the higher dimensional datasets. As evidenced by the MSE comparisons, for the three spatial datasets, the mean predictions of the proposed approach were either more accurate than or at least comparable to those of the state-of-the-art approaches. For the first two higher dimensional datasets, its mean prediction accuracy was comparable to those of the state-of-the-art approaches. For the last dataset, it gave a better MSE performance than the state-of-the-art approaches.

 Second, as evidenced by the NLPD comparisons, the predictive distribution estimated by the proposed approach provides a better fit to test data. This implies that the proposed approach provides better predictive mean and variance estimates when taken together. We designed a simple experiment to investigate this. We used the \texttt{TCO} dataset to compare the degrees of fitness of the predictive distributions from our approach and the PIC, RBCM, and PGP approaches when applied to the test data. In the comparison, the tuning parameters of each compared method were chosen so that the computation time was around 300 seconds, which resulted in approximately the best MSE value for each of the methods. Using these parameters, we calculated the predictive means $\hat{\mu}(x)$ and predictive standard deviations $\hat{\sigma}(x)$ for $x$ in 4,833 test inputs, and we counted the fractions of 4,833 test observations that fell within $\hat{\mu}(x)\pm c\hat{\sigma}(x)$ for different $c$ values, and the fractions were compared with ground truth $P(|X|\le c)$ where $X \sim \mathcal{N}(0,1)$. The fractional numbers closer to the ground truth are better. Table \ref{tbl:posterior} shows the fractional numbers for different $c$ values. The fractions for our method were very close to the ground truth for all choices $c$. The PGP method has much higher fractions than the ground truth, which implies that the PGP tends to overestimate the predictive standard deviation. Both the PIC and the RBCM methods have much lower fractional numbers than the ground truth, which implies that these two local methods significantly underestimate the predictive standard deviation.

Last but not least, the proposed patchwork kriging advances the PGP method by broadening the applicability to higher dimensional datasets, while the PGP method is practically limited to spatial regression with only two input dimensions.
\begin{table}[h]
\centering
\begin{tabular}{|r|r|r|r|r|r|r|}
\hline
$c$                & 0.5 & 1.0 & 1.5 & 2.0 & 2.5 & 3.0 \\
\hline
$P(|X|\le c)$, $X \sim \mathcal{N}(0,1)$ &  0.3829  &  0.6827 &   0.8664 &   0.9545 &   0.9876  &  0.9973\\
Our method &  0.4471  &  0.7366 &   0.8814 &   0.9487 &   0.9766  &  0.9888 \\
PGP &    0.7118  &  0.8850  &  0.9549   & 0.9810  &  0.9905 &   0.9944\\
PIC &    0.0126  &  0.0223  &  0.0362  &  0.0474  &  0.0604  &  0.0741\\
RBCM &    0.1057 &   0.2090  &  0.3095  &  0.3987 &   0.4809 &   0.5535\\
\hline
\end{tabular}
\caption{Percentages of test data ranging in between the estimated predictive mean $\pm c$ the estimated predictive standard deviation. The percentages were compared with $P(|X|\le c)$ where $X$ is a standard normal random variable. The percentage numbers closer to $P(|X|\le c)$ are better. } \label{tbl:posterior}
\end{table}
    
\section{Conclusion} \label{sec:conc}
We presented a new approach to efficiently solve the Gaussian process regression problem for large data. The approach first performs a spatial partitioning of a regression domain into multiple local regions and then assumes a local GP model for each local region. The local GP models are assumed \textit{a priori} independent. However, \textit{a posterior} dependence and related continuity constraints between the local GP models in neighboring regions are achieved by defining an auxiliary process that represents the difference between the local GP models on the boundary of the neighboring regions. By defining zero-valued pseudo observations of the auxiliary process and augmenting the actual data with the pseudo observations, we in essence force the two local GP models to have the same posterior predictive distributions at the collection of boundary points. The proposed idea of enforcing the local models to have the same boundary predictions via pseudo observations is entirely different from that of \citet{park:16}, creating an entirely new framework for patching local GP models. The new approach provides significantly better prediction variance accuracy than the approach of \citet{park:16}, while providing computation efficiency and mean prediction accuracy that are at least comparable and sometimes better. In addition, the spatial partitioning scheme proposed as a part of the new approach makes the new approach applicable for high dimensional regression settings, while the approach of \citet{park:16} is only applicable for one or two dimensional problems. Another advantage of the proposed approach is that its prediction accuracy does not depend strongly on the choice of tuning parameters, so one can simply fine-tune the tuning parameters to minimize the total computation time. Those advantages were numerically demonstrated with six well designed numerical experiments using six real datasets featuring different patterns and dimensions. The new approach has shown better trade-offs between total computation times and prediction accuracy than the approach of \citet{park:16} and other local-based approaches for GP regression. We believe that the proposed patchwork kriging approach is an attractive alternative for large-scale GP regression problems. 

\section*{Acknowledgments}
The authors are thankful for generous support of this work.  Park was
supported in part by the grants from National Science Foundation (CMMI-1334012) and Air
Force Office of Scientific Research (FA9550-13-1-0075 and FA9550-16-1-0110). D. Apley was supported in part by National Science Foundation (CMMI-1537641). 

\section*{Appendix A. Derivation of the predictive mean and variance of the patchwork kriging}
This appendix provides the detailed derivation of the predictive mean and variance in \eqref{eq:posmean} and \eqref{eq:posvar}. From the standard GP modeling result \eqref{eq:joint}, the posterior predictive distribution of $f_{*}^{(k)}$ given $\V{y}$ and the pseudo observations $\V{\delta}$ is Gaussian with mean
\begin{equation*}
\EX[f_{*}^{(k)}|\V{y}, \V{\delta}] = [\V{c}_{*\dataset}^{(k)}, \V{c}_{*, \delta}^{(k)}] \left[\begin{array}{c c}
\V{C}_{\dataset\dataset}   & \V{C}_{\dataset,\delta}  \\
\V{C}_{\delta,\dataset}  & \V{C}_{\delta,\delta}
\end{array}\right]^{-1}\left[\begin{array}{c}
\M{y}  \\
\V{\delta}
\end{array}\right]
\end{equation*}
and variance given below. Using the partitioned matrix inversion formula
\begin{equation} \label{eq:inv_cov}
\left[\begin{array}{c c}
\M{A}  & \V{B}  \\
\V{B}^T  & \M{D}
\end{array}\right]^{-1} = 
\left[\begin{array}{c c}
(\M{A}-\M{B}\M{D}^{-1}\M{B}^T)^{-1}  & -(\M{A}-\M{B}\M{D}^{-1}\M{B}^T)^{-1}\M{B}\M{D}^{-1} \\
-\M{D}^{-1}\M{B}^T(\M{A}-\M{B}\M{D}^{-1}\M{B}^T)^{-1} & (\M{D} - \M{B}^T \M{A}^{-1} \M{B})^{-1}
\end{array}\right],
\end{equation}
we have 
\begin{equation*}
\begin{split}
\EX[f_{*}^{(k)}|\V{y}, \V{\delta}] = &\V{c}_{*\dataset}^{(k)}(\M{C}_{\dataset\dataset}-\M{C}_{\dataset\delta}\V{C}_{\delta,\delta}^{-1}\M{C}_{\dataset\delta}^T)^{-1}\M{y}-\V{c}_{*, \delta}^{(k)}\V{C}_{\delta,\delta}^{-1}\M{C}_{\dataset\delta}^T(\M{C}_{\dataset\dataset}-\M{C}_{\dataset\delta}\V{C}_{\delta,\delta}^{-1}\M{C}_{\dataset\delta}^T)^{-1} \M{y} \\
&-\V{c}_{*\dataset}^{(k)}(\M{C}_{\dataset\dataset}-\M{C}_{\dataset\delta}\V{C}_{\delta,\delta}^{-1}\M{C}_{\dataset\delta}^T)^{-1}\M{C}_{\dataset\delta}\V{C}_{\delta,\delta}^{-1}\V{\delta} + \V{c}_{*, \delta}^{(k)} (\V{C}_{\delta,\delta} - \M{C}_{\dataset\delta}^T \M{C}_{\dataset\dataset}^{-1} \M{C}_{\dataset\delta})^{-1}\V{\delta}.
\end{split}
\end{equation*}
In particular, for $\V{\delta} = \V{0}$, which is the only value for the pseudo observations that we need, we have
\begin{equation*} 
\begin{split} 
\EX[f_{*}^{(k)}|\V{y}, \V{\delta}=\V{0}] = (\V{c}_{*\dataset}^{(k)}- \V{c}_{*\delta}^{(k)}\V{C}_{\delta,\delta}^{-1}\M{C}_{\dataset\delta}^T)(\M{C}_{\dataset\dataset}-\M{C}_{\dataset\delta}\V{C}_{\delta,\delta}^{-1}\M{C}_{\dataset\delta}^T)^{-1}\M{y}.
\end{split}
\end{equation*}
Similarly, the posterior predictive variance of $f_{*}^{(k)}$ given $\V{y}$ and $\V{\delta}$ is 
\begin{equation*}
\VA[f_{*}^{(k)}|\V{y}, \V{\delta}] = c_{**} - [\V{c}_{*\dataset}^{(k)}, \V{c}_{*, \delta}^{(k)}] \left[\begin{array}{c c}
\V{C}_{\dataset\dataset}   & \V{C}_{\dataset,\delta}  \\
\V{C}_{\delta,\dataset}  & \V{C}_{\delta,\delta}
\end{array}\right]^{-1}\left[\begin{array}{c}
\V{c}_{\dataset*}^{(k)}  \\
\V{c}_{\delta*}^{(k)}
\end{array}\right].
\end{equation*}
Applying the matrix inversion result \eqref{eq:inv_cov} to this variance expression, we have 
\begin{equation*} 
\begin{split}
\VA[f_{*}^{(k)}|\V{y}, \V{\delta}] & =  c_{**}- \V{c}_{*\dataset}^{(k)}(\M{C}_{\dataset\dataset}-\M{C}_{\dataset\delta}\V{C}_{\delta,\delta}^{-1}\M{C}_{\dataset\delta}^T)^{-1}\V{c}_{\dataset*}^{(k)} \\
&\quad  +\V{c}_{*, \delta}^{(k)}\V{C}_{\delta,\delta}^{-1}\M{C}_{\dataset\delta}^T(\M{C}_{\dataset\dataset}-\M{C}_{\dataset\delta}\V{C}_{\delta,\delta}^{-1}\M{C}_{\dataset\delta}^T)^{-1} \V{c}_{\dataset*}^{(k)} \\
&\quad  +\V{c}_{*\dataset}^{(k)}(\M{C}_{\dataset\dataset}-\M{C}_{\dataset\delta}\V{C}_{\delta,\delta}^{-1}\M{C}_{\dataset\delta}^T)^{-1}\M{C}_{\dataset\delta}\V{C}_{\delta,\delta}^{-1}\V{c}_{\delta*}^{(k)}\\
&\quad - \V{c}_{*, \delta}^{(k)} (\V{C}_{\delta,\delta} - \M{C}_{\dataset\delta}^T \M{C}_{\dataset\dataset}^{-1} \M{C}_{\dataset\delta})^{-1}\V{c}_{\delta*}^{(k)} \\
& = c_{**}- \V{c}_{*, \delta}^{(k)} \V{C}_{\delta,\delta}^{-1}\V{c}_{\delta*}^{(k)} \\
  & \quad -  (\V{c}_{*\dataset}^{(k)} - \V{c}_{*\delta}^{(k)}\V{C}_{\delta,\delta}^{-1}\M{C}_{\dataset\delta}^T)(\M{C}_{\dataset\dataset}-\M{C}_{\dataset\delta}\V{C}_{\delta,\delta}^{-1}\M{C}_{\dataset\delta}^T)^{-1}(\V{c}_{\dataset*}^{(k)} - \V{C}_{\dataset\delta}\V{C}_{\delta,\delta}^{-1}\M{c}_{\delta*}^{(k)}).
\end{split}
\end{equation*}
\newpage
\section*{Appendix B. Equality of the posterior predictive means and variances of $f_{*}^{(k)}$ and $f_{*}^{(l)}$}
Suppose that $\Gamma_{k,l} \neq \emptyset$, i.e., $\Omega_k$ and $\Omega_l$ are neighboring. This appendix shows
\begin{equation}
\begin{split}
& \EX[f_{*}^{(k)}|\V{y}, \V{\delta}=\V{0}] = \EX[f_{*}^{(l)}|\V{y}, \V{\delta}=\V{0}] \mbox{ for } x_* \in \X{x}_{k,l}, \mbox{ and } \\
& \VA[f_{*}^{(k)}|\V{y}, \V{\delta}] = \VA[f_{*}^{(l)}|\V{y}, \V{\delta}] \mbox{ for } x_* \in \X{x}_{k,l}.
\end{split}
\end{equation}
Let $\V{f}^{(k)}$ denote a column vector of the $f_{*}^{(k)}$ values for $x_* \in \X{x}_{k,l}$ and $\V{f}^{(l)}$ denote a column vector of the $f_{*}^{(l)}$ values for $x_* \in \X{x}_{k,l}$. 
\begin{equation*}
\begin{split}
\EX[\V{f}^{(k)} - \V{f}^{(l)}|\V{y}, \V{\delta}=\V{0}] & = (\M{V}- \M{W}\V{C}_{\delta,\delta}^{-1}\M{C}_{\dataset\delta}^T)(\M{C}_{\dataset\dataset}-\M{C}_{\dataset\delta}\V{C}_{\delta,\delta}^{-1}\M{C}_{\dataset\delta}^T)^{-1}\M{y}.
\end{split}
\end{equation*}
where $\M{V}$ denote the matrix with its $i$th row $\V{c}_{*\dataset}^{(k)}-\V{c}_{*\dataset}^{(l)}$ evaluated for $x_*$ equal to the $i$th entry of $\X{x}_{k,l}$, and let $\M{W}$ denote the matrix with its $i$th row $\V{c}_{*\delta}^{(k)} - \V{c}_{*\delta}^{(l)}$ evaluated for $x_*$ equal to the $i$th entry of $\X{x}_{k,l}$. Note that we can write $\M{V} = \M{\Theta}\M{C}_{\dataset\delta}^T$ and $\M{W} = \M{\Theta}\V{C}_{\delta,\delta}$ for some non-random matrix $\M{\Theta}$. Therefore,
\begin{equation} \label{eq:cont1}
\begin{split}
\EX[\V{f}^{(k)} - \V{f}^{(l)}|\V{y}, \V{\delta}=\V{0}] & = \M{\Theta}(\M{C}_{\dataset\delta}^T- \V{C}_{\delta,\delta}\V{C}_{\delta,\delta}^{-1}\M{C}_{\dataset\delta}^T)(\M{C}_{\dataset\dataset}-\M{C}_{\dataset\delta}\V{C}_{\delta,\delta}^{-1}\M{C}_{\dataset\delta}^T)^{-1}\M{y}\\
& = \V{0}.
\end{split}
\end{equation}
Similarly, we have
\begin{equation} \label{eq:cont2}
\begin{split}
\VA[\V{f}^{(k)}|\V{y}, \V{\delta}] - \VA[\V{f}^{(l)}|\V{y}, \V{\delta}] & = \M{\Theta}(\M{C}_{\dataset\delta}^T- \V{C}_{\delta,\delta}\V{C}_{\delta,\delta}^{-1}\M{C}_{\dataset\delta}^T)(\M{C}_{\dataset\dataset}-\M{C}_{\dataset\delta}\V{C}_{\delta,\delta}^{-1}\M{C}_{\dataset\delta}^T)^{-1}\M{Z}\\
& = \V{0},
\end{split}
\end{equation}
where $\M{Z}$ denote the matrix with its $j$th column $\V{c}_{\dataset*}^{(k)}+\V{c}_{\dataset*}^{(l)}$ evaluated for $x_*$ equal to the $j$th entry of $\X{x}_{k,l}$. 

\newpage
\section*{Appendix C. Comparison of the patchwork kriging and the full GP for the simulated cases in Section \ref{sec:simstudy}}
\begin{figure}[ht!]
    \centering
            \includegraphics[width=\textwidth]{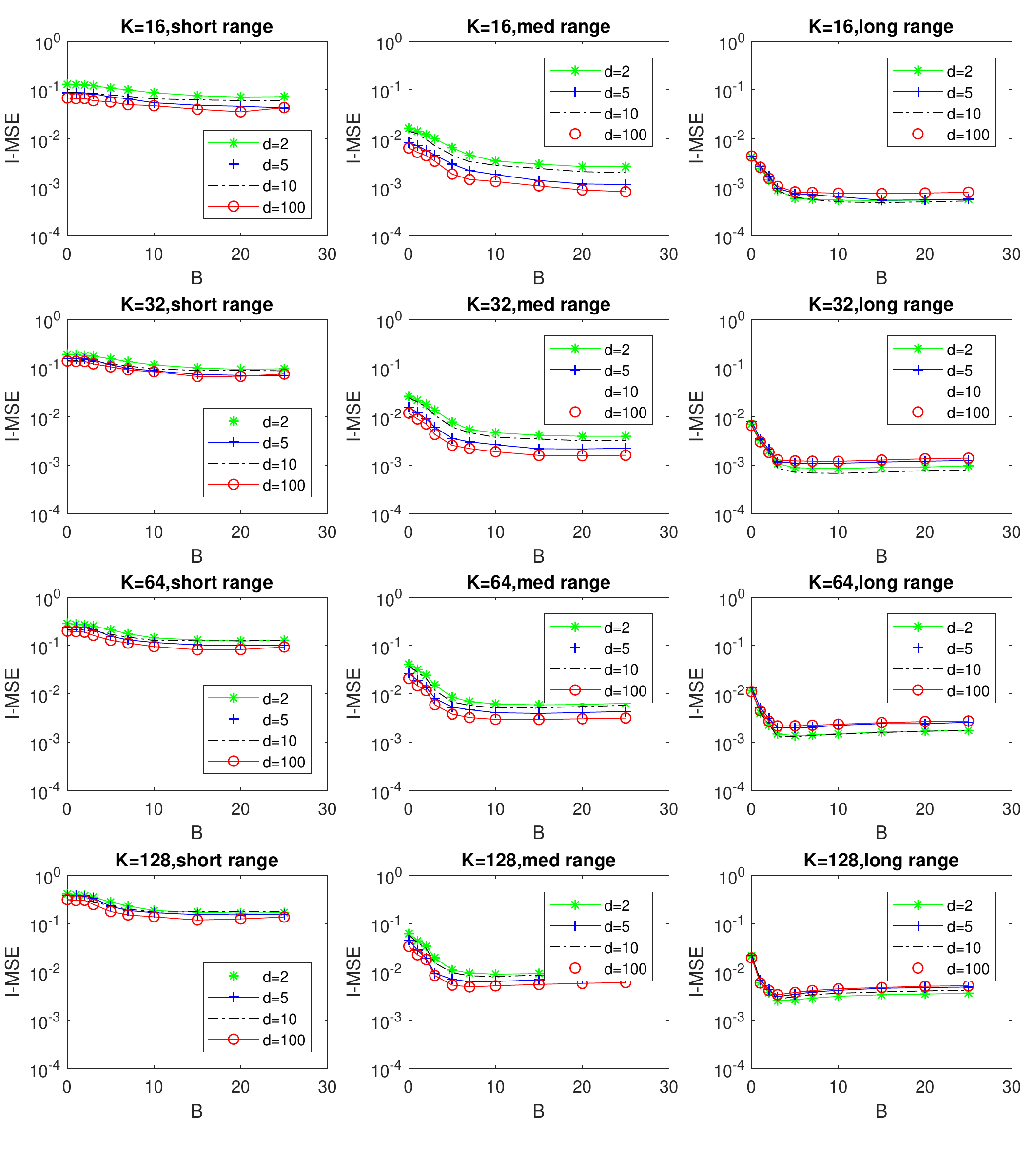}
    \caption{Summary of Interior Mean Squared Error for Predictive Mean (I-MSE)}
    \label{fig:I-MSE}
\end{figure}

\begin{figure}[ht!]
    \centering
            \includegraphics[width=\textwidth]{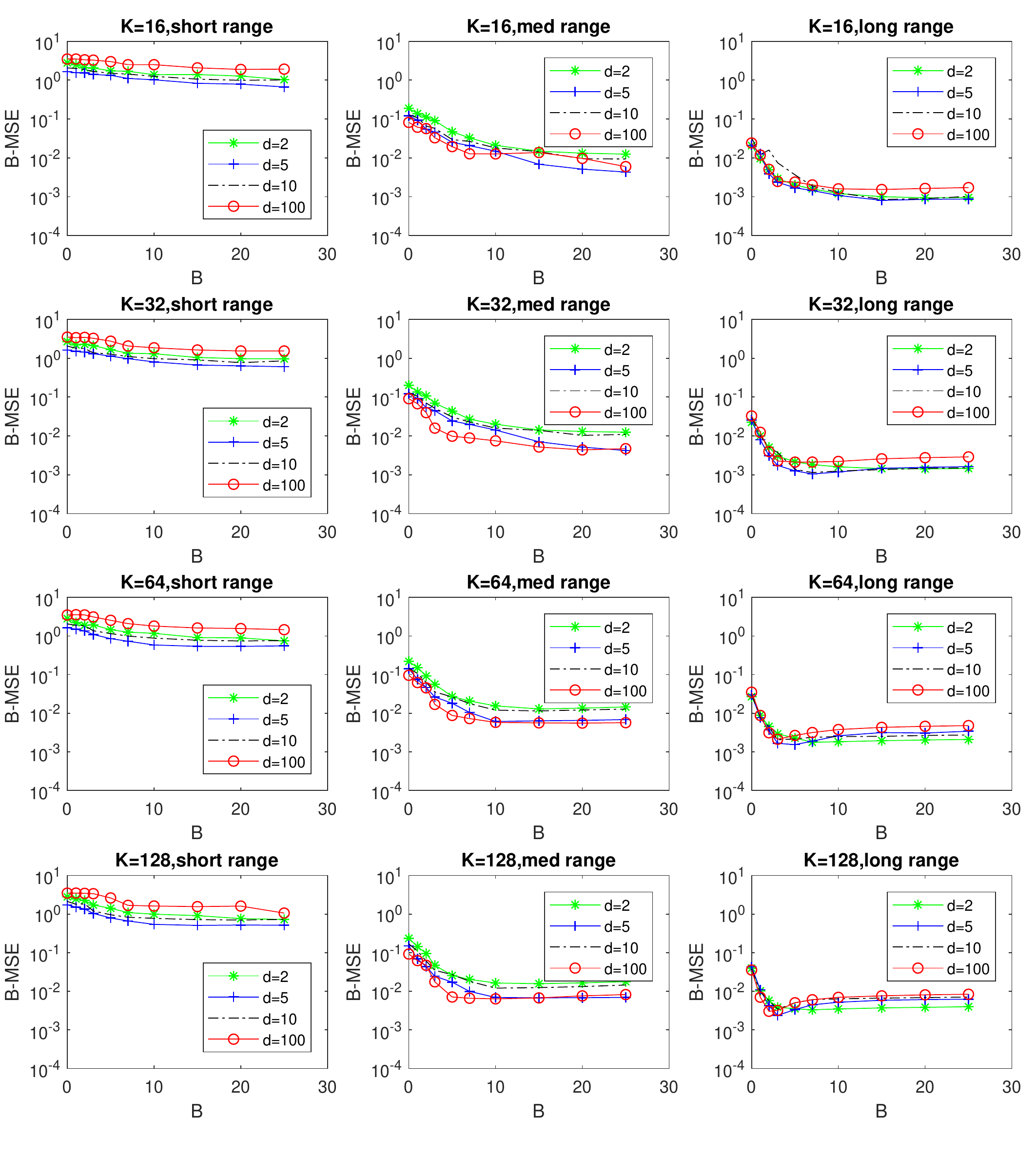}
    \caption{Summary of Boundary Mean Squared Error for Predictive Mean (B-MSE)}
    \label{fig:B-MSE}
\end{figure}

\begin{figure}[ht!]
    \centering
            \includegraphics[width=\textwidth]{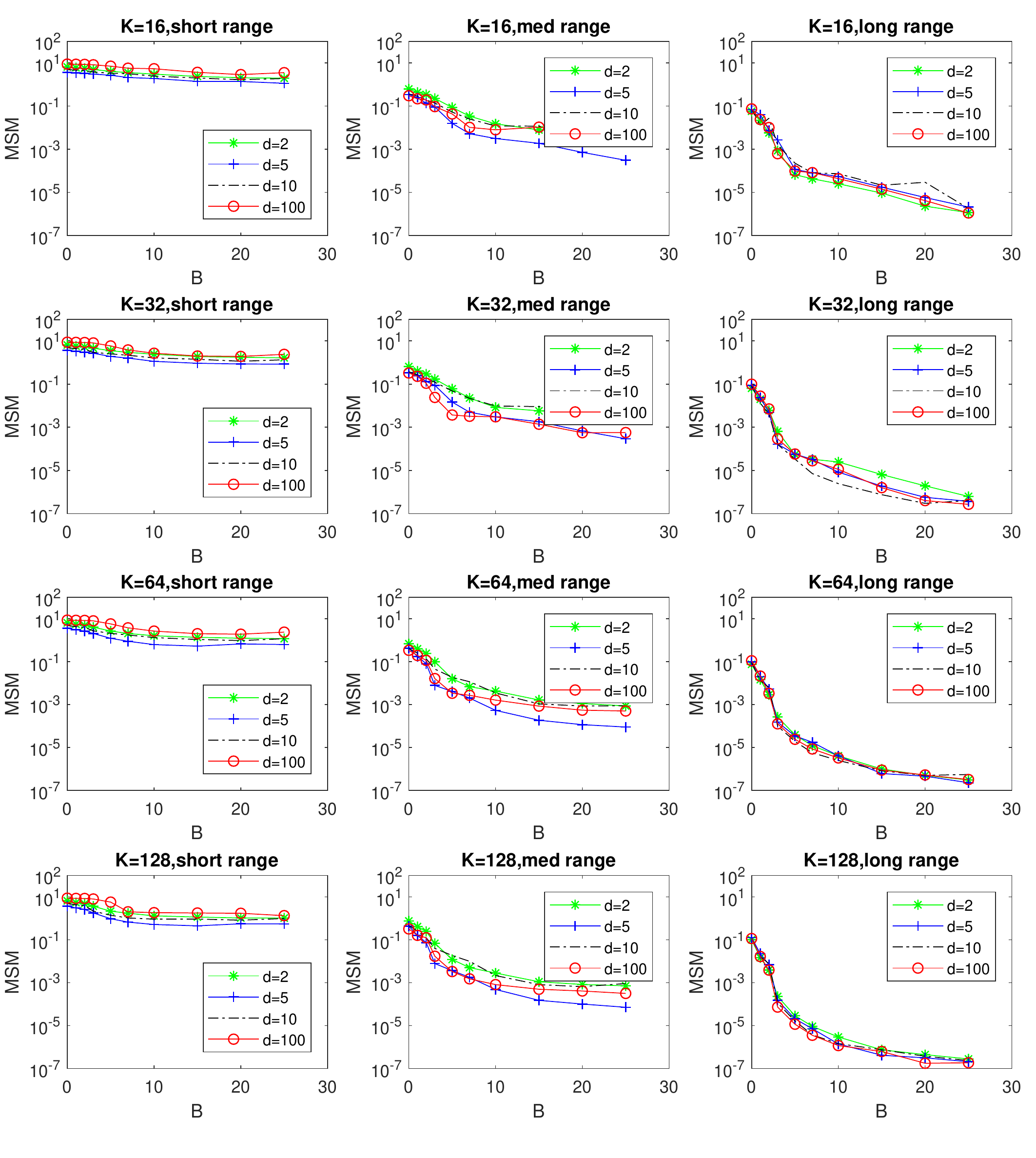}
    \caption{Summary of Mean Squared Mismatch for Predictive Mean (MSM)}
    \label{fig:MSM}
\end{figure}

\begin{figure}[ht!]
    \centering
            \includegraphics[width=\textwidth]{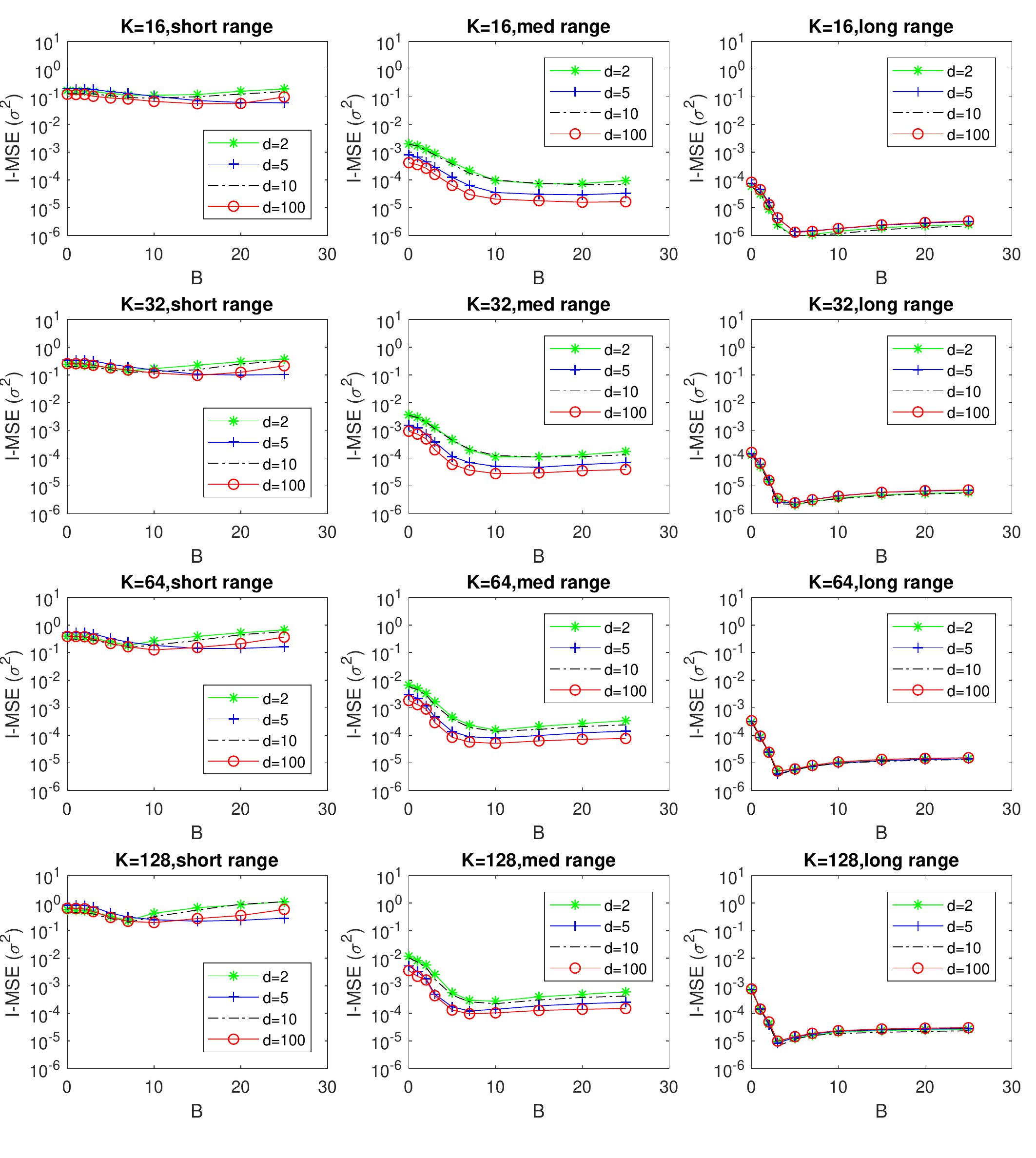}
    \caption{Summary of Interior Mean Squared Error for Predictive Variance (I-MSE($\sigma^2$))}
    \label{fig:I-MSE-sig}
\end{figure}

\begin{figure}[ht!]
    \centering
            \includegraphics[width=\textwidth]{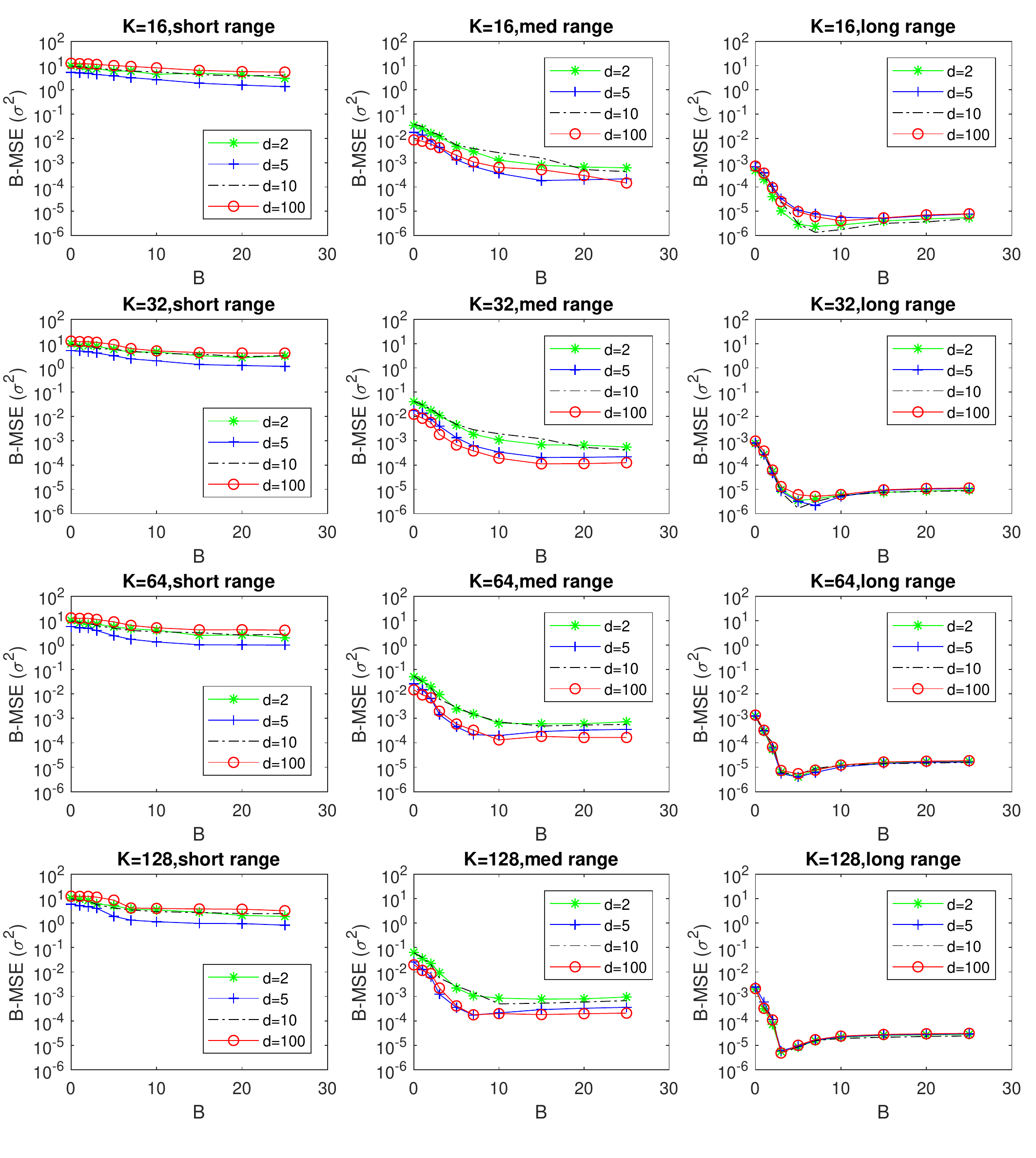}
    \caption{Summary of Boundary Mean Squared Error for Predictive Variance (B-MSE($\sigma^2$))}
    \label{fig:B-MSE-sig}
\end{figure}

\begin{figure}[ht!]
    \centering
            \includegraphics[width=\textwidth]{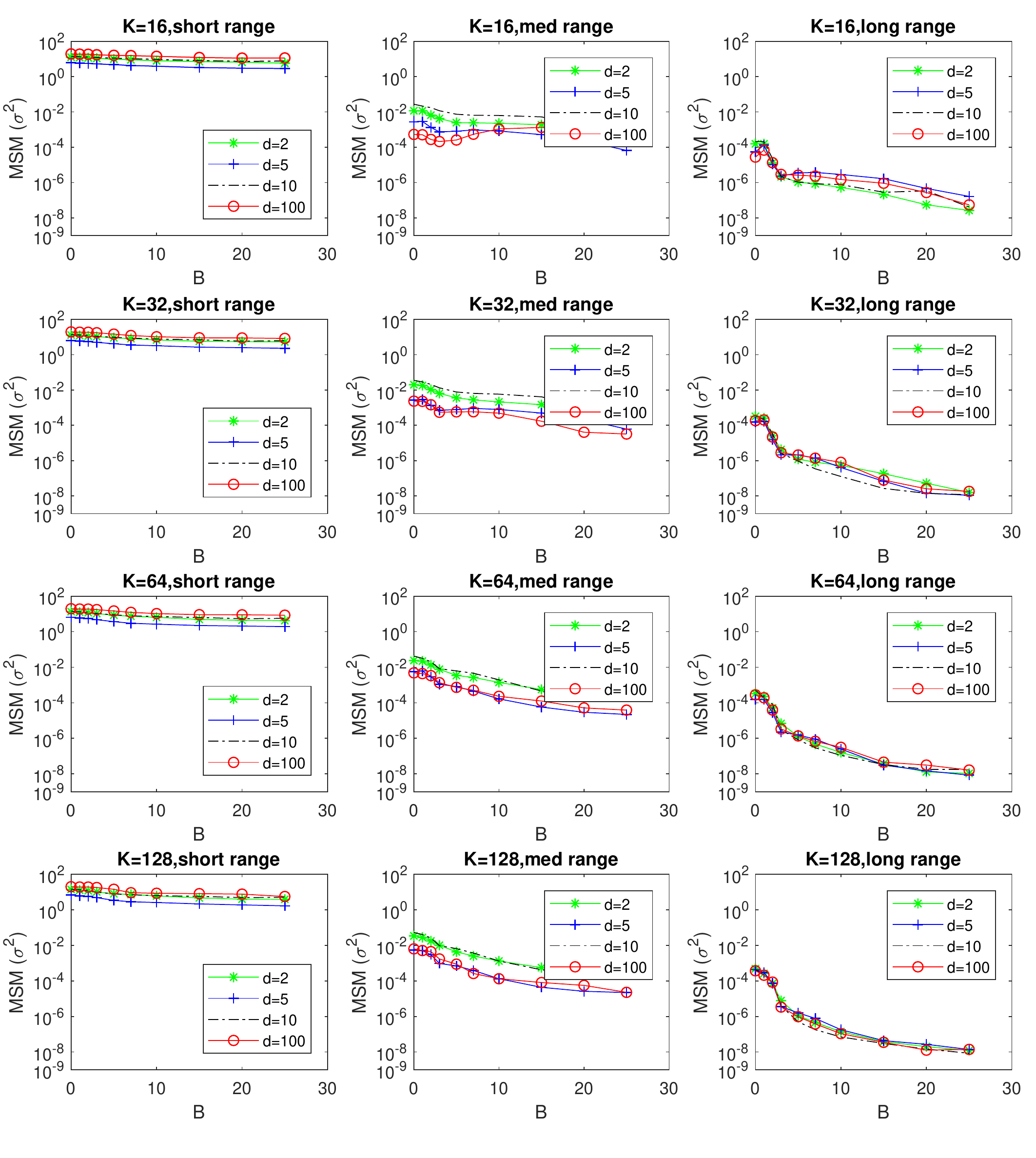}
    \caption{Summary of Mean Squared Mismatch for Predictive Variance (MSM($\sigma^2$))}
    \label{fig:MSM-sig}
\end{figure}

This appendix presents the full comparison data that are summarized in Section \ref{sec:simstudy}. Figures \ref{fig:I-MSE}, \ref{fig:B-MSE} and \ref{fig:MSM} shows the comparison in a posterior mean prediction, and Figures \ref{fig:I-MSE-sig}, \ref{fig:B-MSE-sig} and \ref{fig:MSM-sig} shows the comparison in a posterior variance prediction
\newpage
\newpage
\newpage
\newpage
\newpage
\newpage

\section*{Appendix D. Extra real data study}
This appendix includes the numerical comparison with an extra real dataset, which were not in the main text. The dataset, \texttt{TCO}, contains 48,331 observations collected by the
NIMBUS-7/TOMS satellite, which measures the total column of ozone over the globe on Oct 1 1988. This dataset has two input dimensions that represent a spatial location of the measurement, and the inputs of the data are densely distributed over a rectangular domain. For patchwork kriging, we varied $B\in\{5,7\}$ and  $K \in \{64,128,256,512\}$. The prediction accuracy of the PGP did not depend on the number of local regions $K$, so we fixed $K=145$, while the number of finite element meshes per local region was varied from 5 to 40 with step size 5. For RBCM, we varied the number of local experts $K \in \{100, 150, 200, 250, 300, 600\}$. For PIC, $K$ was varied over $\{100, 150, 200, 250, 300, 400\}$, and the total number of pseudo inputs was also varied over $\{30,50,70,80,100,150,200,250,300\}$. For the GMRF, following the suggestion by the GMRF's authors, we used the voronoi-tessellation of training points for mesh generation. 

\begin{figure}[t]
    \centering
            \includegraphics[width=0.9\textwidth]{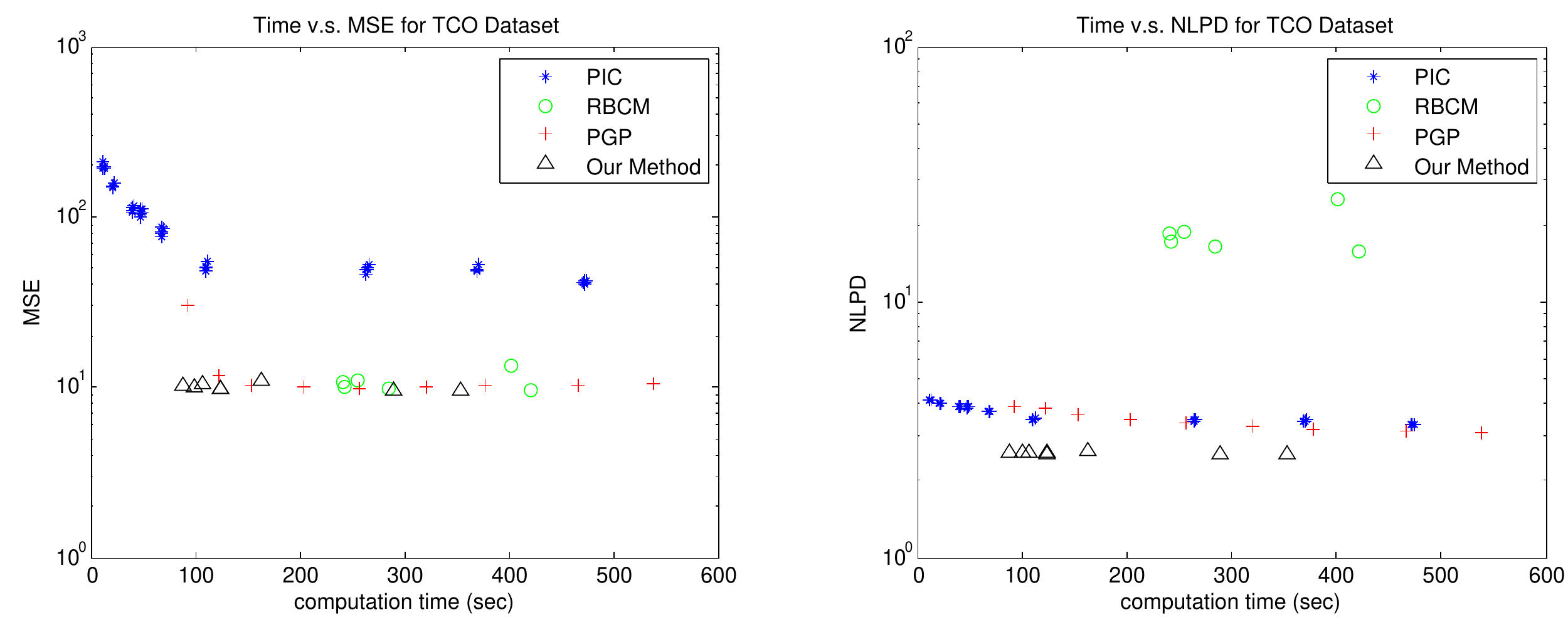}
            \includegraphics[width=0.9\textwidth]{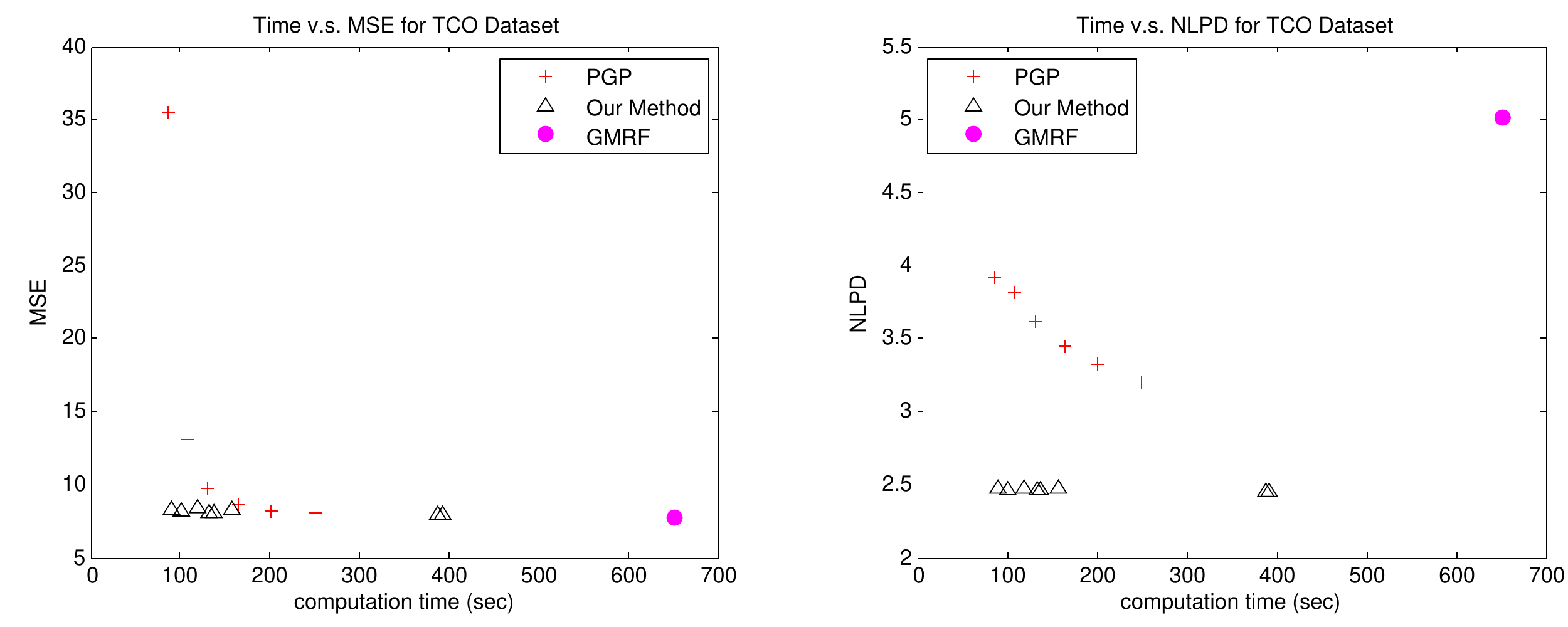}
    \caption{Prediction accuracy versus total computation time for the \texttt{TCO} data. Top panel: a squared exponential covariance function was used. Bottom panel: an exponential covariance function was used. In the top panel, eight triangles are supposed to show up. However, two of the eight triangles are very closely located, so it looks like that there are only 7 triangles.}
    \label{fig:result_tco}
\end{figure}

Figure \ref{fig:result_tco} summarizes the performance results. The two panels in the top row compare the proposed approach to the PGP, RBCM and PIC approaches when a square exponential covariance function is used. The MSE plot shows the mean square error of the mean predictions for the test data. The proposed approach and the PGP approach had better MSE than the RBCM and PIC approaches, and the proposed approach also had better MSE than the PGP approach at the 100 second computation time. The NLPD plot shows the degree of fitness of the estimated predictive distribution to the test data, which is affected by both of the mean prediction and the variance prediction. The proposed approach uniformly outperforms the other methods, including the PGP. The two panels in the bottom row compare the proposed approach to the PGP and the GMRF approaches when an exponential covariance function is used. The PGP and the proposed approaches had comparable MSEs at larger computation times, whereas the proposed approach had much smaller MSE at the lower computation times, and the proposed approach outperformed the PGP and the GMRF in terms of NLPD. The GMRF had longer computation time than the PGP and the proposed method.

\vskip 0.2in
\bibliography{patchkrig}

\end{document}